\documentclass[journal]{IEEEtran}
\ifCLASSINFOpdf
\else
\fi

\usepackage[sorting=none]{biblatex} 
\usepackage{amsmath,graphicx,svg}
\usepackage{adjustbox,subcaption, booktabs,amssymb,hyperref, multirow}

\usepackage{adjustbox}
\begin{document}
%

\title{Leveraging Vision-Language Models \\ to Select Trustworthy Super-Resolution Samples \\ Generated by Diffusion Models}

%
%
%

\author{Cansu~Korkmaz,~\IEEEmembership{Member,~IEEE,}
        A. Murat~Tekalp,~\IEEEmembership{Fellow,~IEEE}
        and~Zafer~Do\u{g}an,~\IEEEmembership{Member,~IEEE}

\thanks{Copyright © 2025 IEEE. Personal use of this material is permitted. However, permission to use this material for any other purposes must be obtained from the IEEE by sending an email to pubs-permissions@ieee.org. Authors are with the Department of Electrical and Electronics Engineering and KUIS AI Center, Koç University, 34450 Istanbul, Turkey. C.K. was supported by a KUIS AI Center Fellowship. A.M.T. was supported by TUBITAK 2247-A Award No. 120C156 and by the Turkish Academy of Sciences (TUBA). Z.D. was supported by TUBITAK 2232 International Fellowship for Outstanding Researchers Award No. 118C337.}}

%
%

\markboth{IEEE TRANSACTIONS ON CIRCUITS AND SYSTEMS FOR VIDEO TECHNOLOGY,~VOL.~XX, NO.~X, XXXX~2025}%
{Shell \MakeLowercase{\textit{et al.}}: Bare Demo of IEEEtran.cls for IEEE Journals}
%



\maketitle

\begin{abstract}
Super-resolution (SR) is an ill-posed inverse problem with many feasible solutions that are consistent with a given low-resolution image. On one hand, regressive SR models aim to balance fidelity and perceptual quality to yield a single solution; but this trade-off often leads to artifacts that introduce ambiguity in information-critical applications such as identifying digits or letters. On the other hand, diffusion models generate a diverse set of SR images; but now selecting the most trustworthy solution out of this set becomes a challenge. This paper introduces a robust, automated framework for identifying the most trustworthy SR sample from a diffusion-generated set by leveraging the semantic reasoning capabilities of vision-language models (VLMs). Specifically, VLMs such as BLIP-2, GPT-4o, and their variants are prompted with structured queries to evaluate semantic correctness, visual quality, and the presence of artifacts. The top-ranked SR candidates are then ensembled to yield a single trustworthy output in a cost-effective manner. To rigorously assess the validity of VLM-selected samples, we propose a novel Trustworthiness Score (TWS)—a hybrid metric that quantifies SR reliability based on three complementary components: semantic similarity using CLIP embeddings, structural integrity via SSIM on edge maps, and artifact sensitivity measured through a multi-level wavelet decomposition. We empirically demonstrate that TWS correlates strongly with human preference in both ambiguous and natural images, and that VLM-guided selections consistently yield high TWS values. Compared to conventional metrics like PSNR, LPIPS, and DISTS—which fail to reflect information fidelity—our approach offers a principled, scalable, and generalizable solution for navigating the uncertainty of the diffusion SR space. By aligning model outputs with human expectations and semantic correctness, this work sets a new benchmark for trustworthiness in generative SR tasks.
\end{abstract}


%
\IEEEpeerreviewmaketitle

\section{Introduction}
%
%
%
%

\begin{figure}[!t]
\centering
\begin{subfigure}{0.155\textwidth}
\centering
\captionsetup{justification=centering}
    \begin{subfigure}{\textwidth}
        \includegraphics[width=\textwidth]{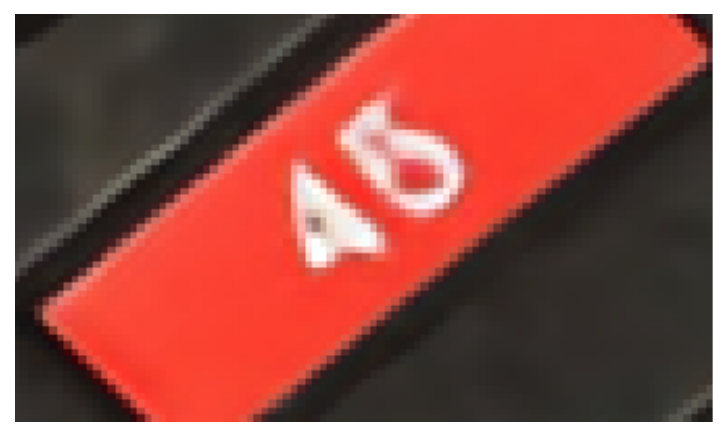}
    \end{subfigure} \vspace{-16pt}
    \caption*{EDSR  \cite{EDSR2017}\\ (27.17 / 0.213) \\ 0.2804}
    \begin{subfigure}{\textwidth}
        \includegraphics[width=\textwidth]{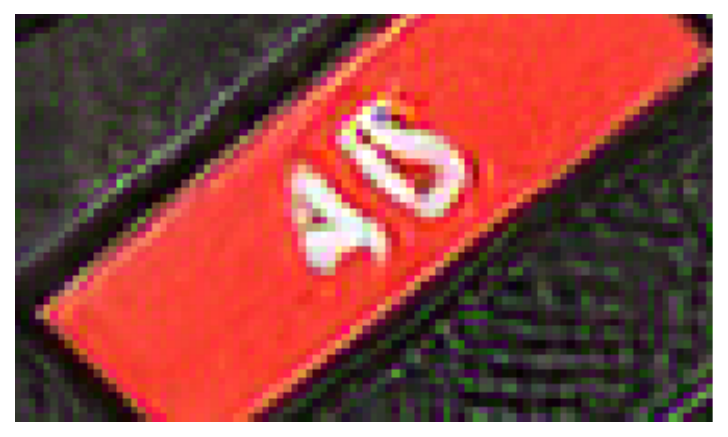}
    \end{subfigure} \vspace{-16pt}
    \caption*{ ESRGAN+ \cite{esrganplus} \\ (22.78 / 0.197) \\ 0.1419}
    \begin{subfigure}{\textwidth}
        \includegraphics[width=\textwidth]{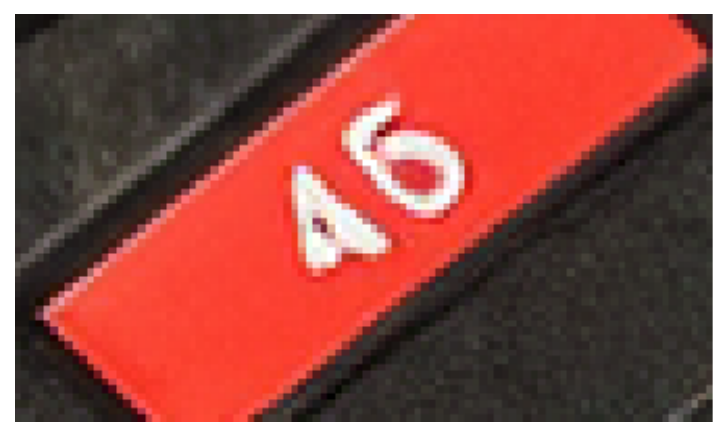}
    \end{subfigure} \vspace{-16pt}
    \caption*{SROOE \cite{srooe_Park_2023_CVPR} \\ (26.67 / 0.184) \\ 0.2489}
    \begin{subfigure}{\textwidth}
        \includegraphics[width=\textwidth]{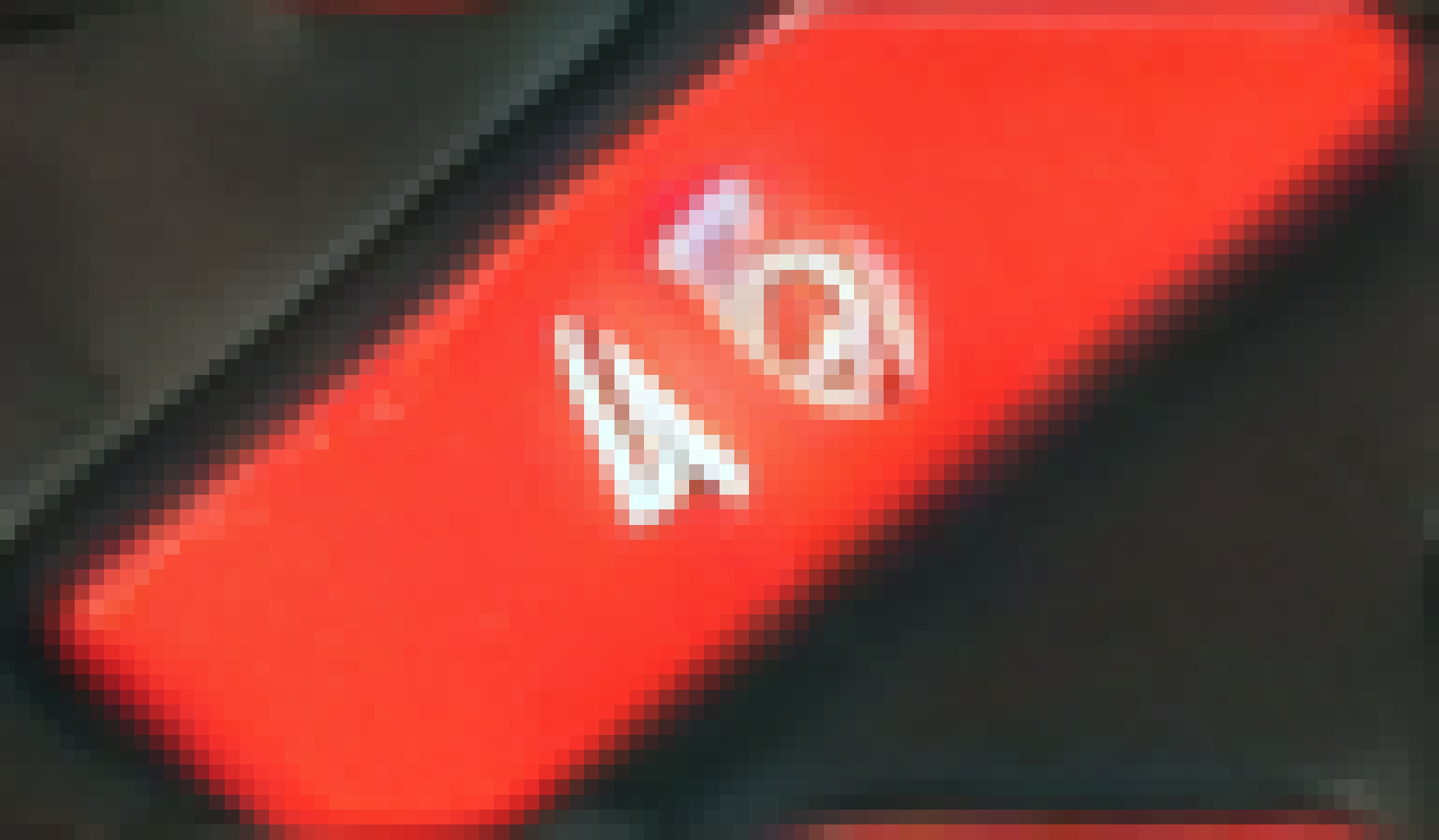}
    \end{subfigure} \vspace{-16pt}
    \caption*{PASD \cite{yang2023pasd} \\ (23.98 / 0.292) \\ 0.2740}
\end{subfigure}
\begin{subfigure}{0.155\textwidth}
\centering
\captionsetup{justification=centering}
    \begin{subfigure}{\textwidth}
        \includegraphics[width=\textwidth]{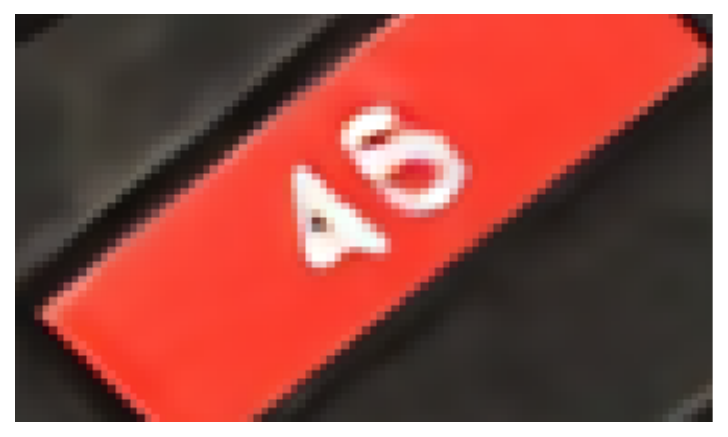}
    \end{subfigure} \vspace{-16pt}
    \caption*{RRDB \cite{wang2018esrgan} \\ (26.46 / 0.223) \\ 0.2729}
    \begin{subfigure}{\textwidth}
        \includegraphics[width=\textwidth]{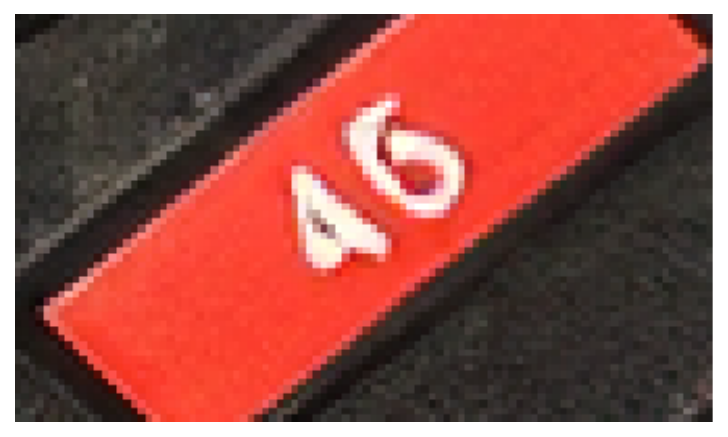}
    \end{subfigure} \vspace{-16pt}
    \caption*{SPSR \cite{ma_SPSR} \\ (25.82 / 0.184) \\ 0.2295}
    \begin{subfigure}{\textwidth}
        \includegraphics[width=\textwidth]{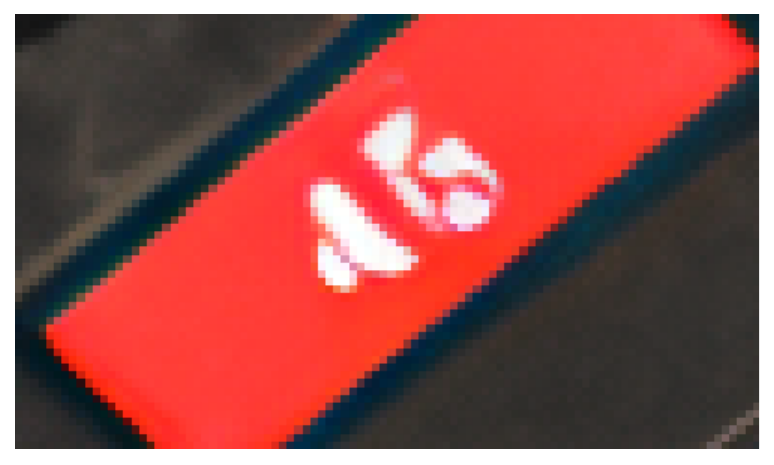}
    \end{subfigure} \vspace{-16pt}
    \caption*{LDM \cite{LDM_rombach2022high} \\ (24.71 / 0.246) \\ 0.2945}
    \begin{subfigure}{\textwidth}
        \includegraphics[width=\textwidth]{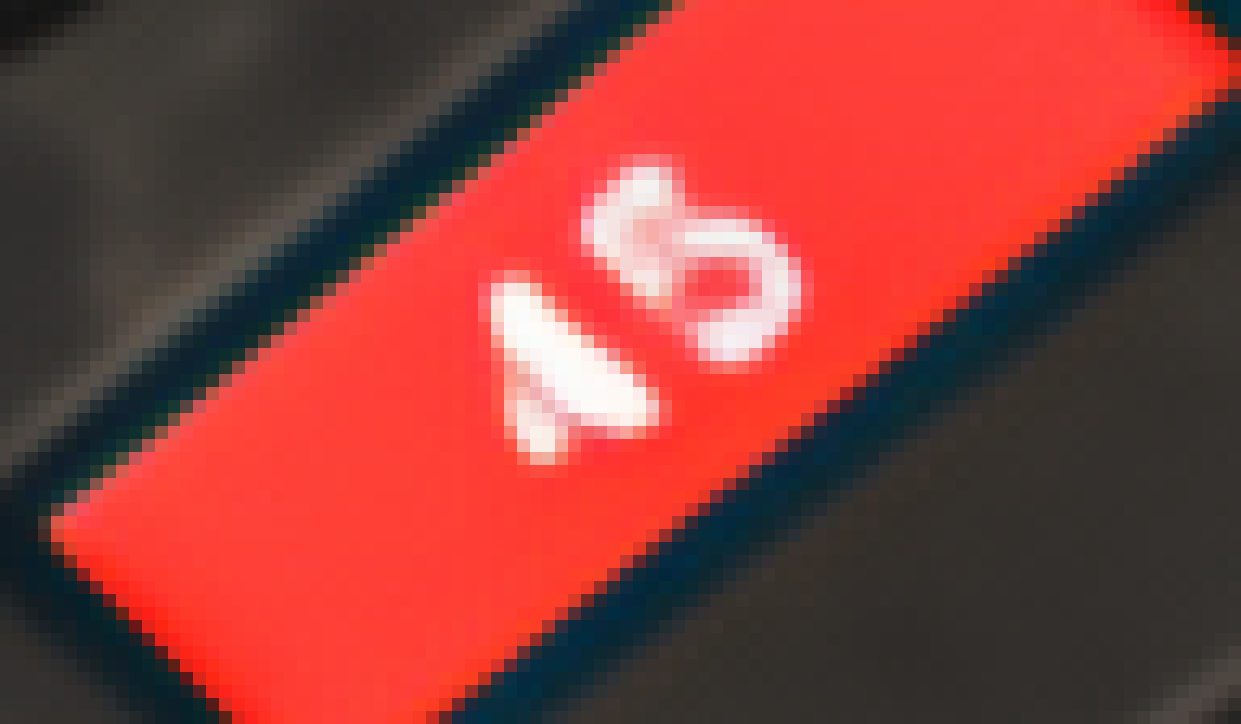}
    \end{subfigure} \vspace{-16pt}
    \caption*{LDM-VLM (Ours) \\ (25.54 / 0.270) \\ 0.3050}
\end{subfigure}
\begin{subfigure}{0.155\textwidth}
\centering
\captionsetup{justification=centering}
     \begin{subfigure}{\textwidth}
        \includegraphics[width=\textwidth]{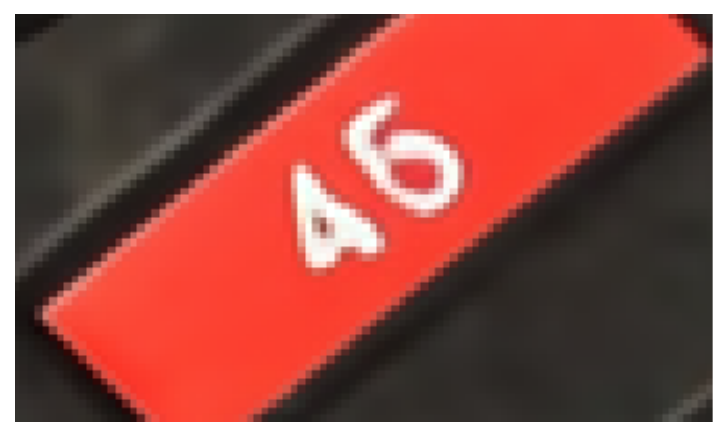}
    \end{subfigure} \vspace{-16pt}
    \caption*{ HAT \cite{hat_chen2023activating} \\ (28.03 / 0.217) \\ 0.2786}
    \begin{subfigure}{\textwidth}
        \includegraphics[width=\textwidth]{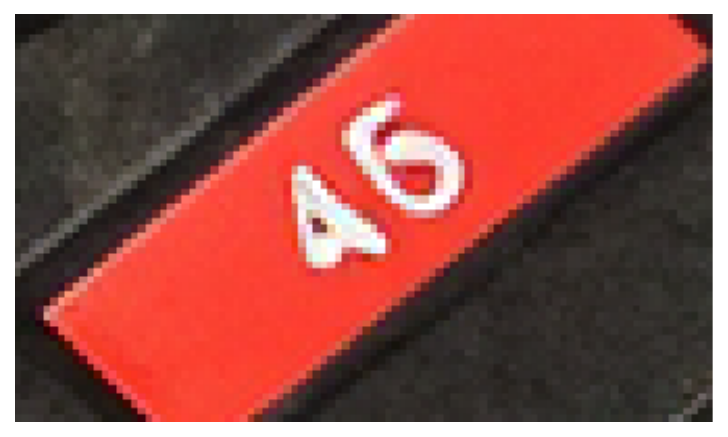}
    \end{subfigure} \vspace{-16pt}
    \caption*{ LDL \cite{details_or_artifacts} \\ (26.67 / 0.194) \\ 0.2551}
    \begin{subfigure}{\textwidth}
        \includegraphics[width=\textwidth]{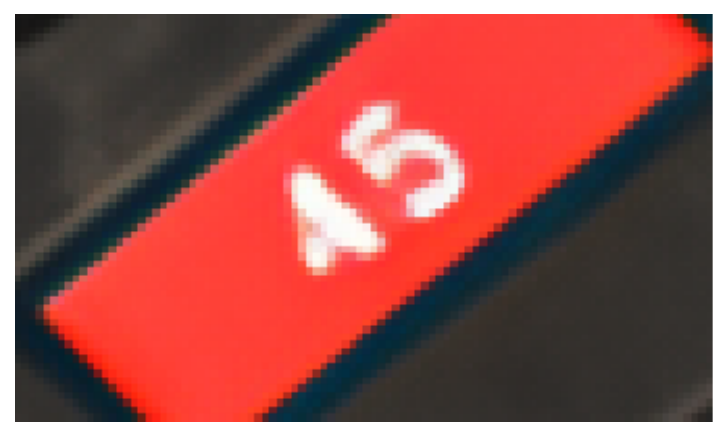}
    \end{subfigure} \vspace{-16pt}
    \caption*{LDM-HS \cite{korkmaz2024_icip} \\ (26.69 / 0.193) \\ 0.3033}
    \begin{subfigure}{\textwidth}
        \includegraphics[width=\textwidth]{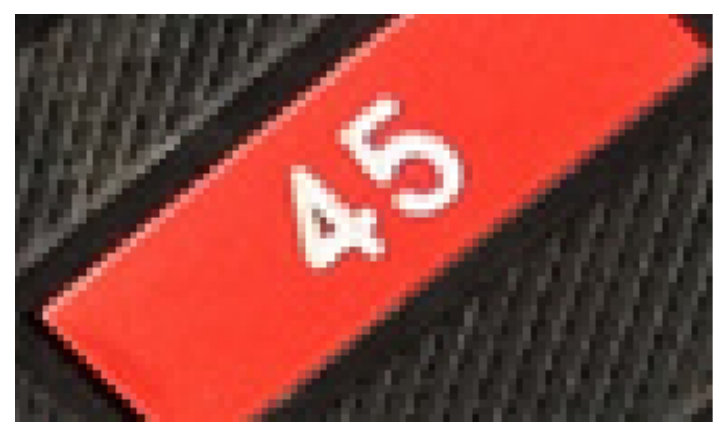}
    \end{subfigure} \vspace{-16pt}
    \caption*{HR \\ (PSNR$\uparrow$ / DISTS$\downarrow$) \\ TWS$\uparrow$}
\end{subfigure} 
\caption{Ambiguity in SR: Results of the state-of-the-art models for $\times$ 4 SR on a~crop from img-6 of Urban100 dataset \cite{urban100_cite}. SOTA methods reconstruct ``5" as ``6", whereas the opening in the lower part of ``5" is visible in our results confirming  our proposed strategy helps resolve ambiguity to provide more reliable solutions. Note quantitative scores such as PSNR, DISTS are not good indicators~of~information~trustworthiness. In contrast, our Trustworthiness Score (TWS) reflects the advantage of LDM-VLM indicating a more reliable and semantically faithful reconstruction.}
\label{fig:first_img} 
\end{figure}

\IEEEPARstart{S}{ingle} image super-resolution (SR) is fundamentally an ill-posed inverse problem, wherein multiple plausible high-resolution (HR) images can be generated from a single low-resolution (LR) image~\cite{tekalp2022}. This inherent ambiguity poses significant challenges in information-critical applications that require precise outputs from SR reconstructions, such as digit or letter recognition. Early supervised deep learning approaches~\cite{dong2015image, ledig2017photorealistic, EDSR2017, RCAN2018, wang2018esrgan} framed SR as a regularization problem using paired LR-HR data. These methods relied on image priors to mitigate ambiguity and produce a single output. However, this approach often results in high-frequency artifacts that can lead to erroneous conclusions in downstream tasks, as illustrated in Figure~\ref{fig:first_img}. 

\begin{figure*}
\centering
\includegraphics[width=\linewidth]{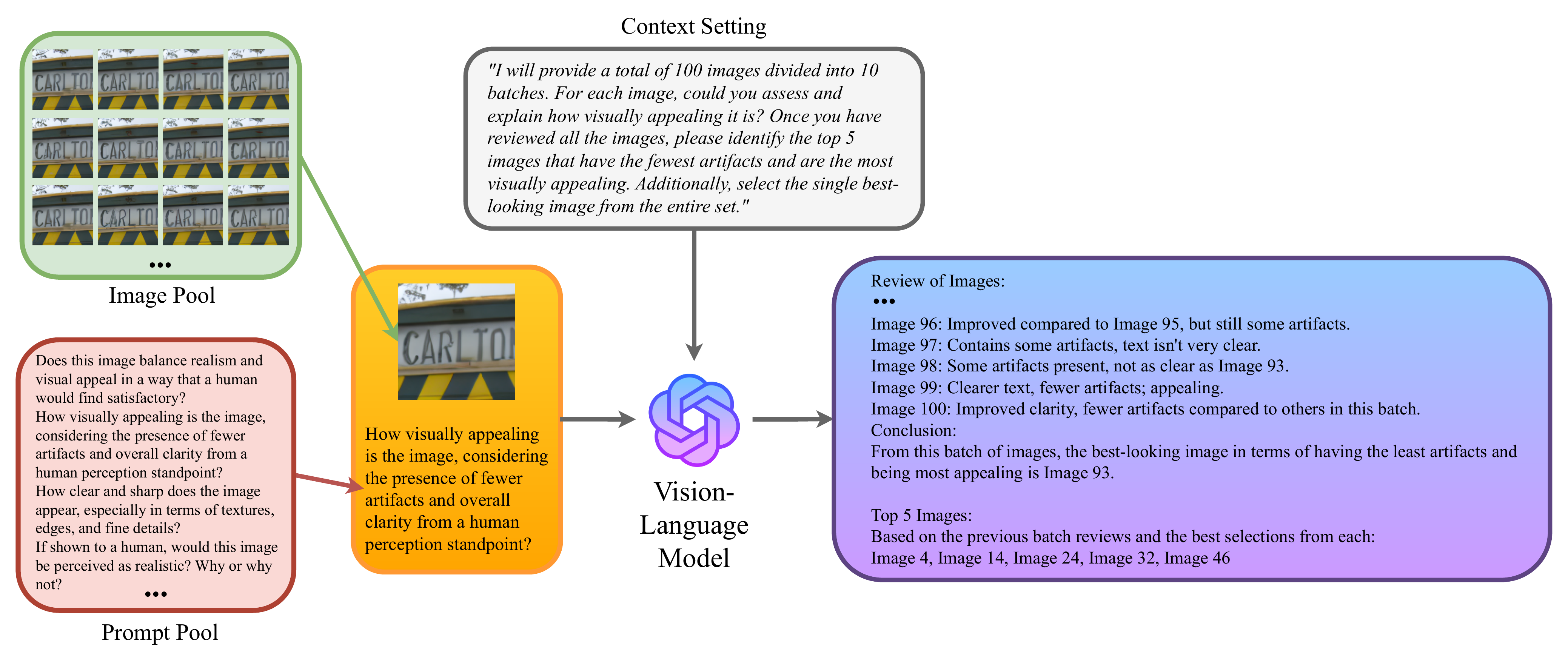} \vspace{-15pt}
\caption{Vision-language models (VLMs) enable automatic selection of reliable SR samples out of a pool of samples generated by diffusion SR models via proper context setting and prompts from a pool providing accuracy comparable to human evaluators.}
\label{fig:gpt4_depiction}
\end{figure*}

More recent approaches~\cite{jo2021srflowda, SR3_saharia2022image, korkmaz2022perception, diff_mixture_of_experts_luo2023image} proposed stochastic solutions that aim to generate a diverse set of SR images from an LR image, to effectively cover the solution space through one-to-many mappings. 
Diffusion models (DMs) \cite{LDM_rombach2022high, diff_chung2022improving, diff_implicit_chen2021learning, diff_mixture_of_experts_luo2023image, lora_diff_realsr_wu2024one, lora_diff_realsr_wu2024seesr, yang2023pasd, wang2024exploiting} have been shown to generate diverse, high-quality SR outputs by sampling the~conditional distribution of plausible HR images given an LR input. While DMs succeed in creating a broad variety of visually appealing solutions, they introduce a new~challenge: how to determine a single trustworthy solution out of many plausible samples when the task demands interpretation of specific information—such as identifying digits or letters. In such tasks, photorealism and perceptual quality are secondary to the~accuracy of the~information conveyed by the SR image.

Traditionally, the performance of SR algorithms is evaluated by means of pixel-wise fidelity metrics, such as Peak Signal-to-Noise Ratio (PSNR), Structural Similarity Index (SSIM), and/or perceptual quality motivated feature-level metrics, such as Learned Perceptual Image Patch Similarity (LPIPS)~\cite{lpips}, Deep Image Structure and Texture Similarity (DISTS)~\cite{dists}, and Fréchet Inception Distance (FID) \cite{heusel2017gans}. We observed that all of these quantitative measures have limited value in assessing trustworthiness of the information content of an image, mainly because fidelity measures provide an average of pixel-wise differences over an image rather than focusing on critical information on a specific region of the image, while feature-based measures evaluate naturalness of the result~rather~than~its~fidelity. Therefore, one cannot rely solely on classical quantitative measures to evaluate the trustworthiness of information, e.g., identifying digits or letters, inferred from ambiguous SR outputs.
As a result, there is a pressing need for alternative evaluation strategies to better assess the trustworthiness of SR results, especially when extracting accurate information is essential.

To address this gap, we propose a novel framework for selecting and verifying trustworthy SR outputs from diffusion models. Our method leverages vision-language models~(VLMs), including BLIP-2~\cite{blip2_li2023blip}, GPT-4o~\cite{gpt4_achiam2023gpt}, and their variants, to assess the semantic and perceptual quality of each SR candidate. As illustrated in Figure~\ref{fig:gpt4_depiction}, through prompt-based querying (e.g.,``What is the digit?”, ``Are there visible artifacts?”, ``How appealing is the image?”), VLMs act as automated evaluators that identify samples preserving both visual quality and critical information. The top-ranked candidates are ensembled to produce a single reliable SR image in a cost-efficient and scalable manner.

In parallel, we introduce a novel Trustworthiness Score~(TWS)—a hybrid metric to quantify the reliability of SR outputs across three complementary dimensions. TWS integrates (1) semantic similarity measured via CLIP embeddings, (2)~edge consistency evaluated using SSIM, and (3) artifact penalization through multi-level wavelet decomposition. Wavelet differences are computed across fine-to-coarse scales and normalized to bring them in line with the range of semantic and structural scores. This balanced formulation allows TWS to serve as both a verification tool for VLM selections and a general-purpose trustworthiness estimator, particularly in the~absence of ground truth.

To evaluate the accuracy of VLMs in assessing the trustworthiness of SR images, we also employ a human-in-the-loop approach. Human participants are asked to evaluate samples generated by diffusion models—identifying the number or letter in information-critical tasks, and selecting images with fewer artifacts in natural scenes. We compare their selections with those made by VLMs and show that VLM-based choices not only align closely with human judgment, but also provide high TWS. This demonstrates that VLMs provide a scalable alternative to trustworthy visual decision making that align with human decision-making 
with a high degree of accuracy.

\noindent Our main contributions can be summarized as follows: \\
1. We introduce a scalable and automated VLM-based evaluation framework for selecting trustworthy SR outputs from diffusion models. \\
2. We propose TWS, a hybrid metric that combines CLIP-based semantic similarity, edge-aware SSIM, and wavelet-based artifact analysis to quantify trustworthiness in SR images. \\
3. We demonstrate that VLM-selected SR images not only match human evaluations but also consistently achieve high TWS values, confirming their reliability across various domains. \\

We elaborate on related works and the novelty of this paper in Section~\ref{related}. Our method is detailed in Section~\ref{method}, and experimental results—including prompt robustness and human alignment analyses—are presented in Section~\ref{exp}. Section~\ref{conc} concludes the paper.

\begin{figure*}[t]
\centering
\includegraphics[width=\linewidth]{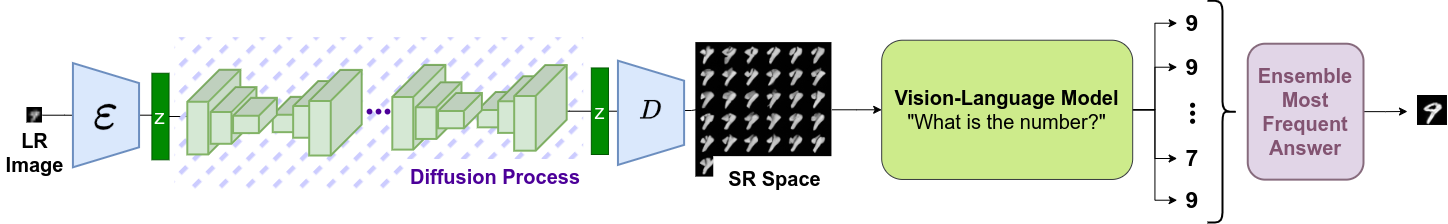}
\caption{The block diagram of our proposed trustworthy SR framework, consisting of diffusion SR sample space generation, automated reliable sample selection by VLMs, and ensembling of selected samples. VLMs evaluate each diffusion SR sample through natural language queries, allowing us to ensemble the most frequently selected samples by the model. }
\label{fig:arch}
\end{figure*}

\section{Related Work and Contributions}
\label{related}

\subsection{One-to-One SR Inference} 

Many prominent CNN-based SR models \cite{ledig2017photorealistic, zhang2018residual, liang2021swinir}, including 
EDSR \cite{EDSR2017}, RRDB~\cite{wang2018esrgan}, RCAN \cite{RCAN2018} and HAT \cite{hat_chen2023activating}, are one-to-one regressive mappings from LR to HR images trained by $l_1$ or $l_2$ pixel reconstruction losses. Although these models achieve high fidelity as measured by PSNR, they frequently produce significant artifacts that exacerbate the ambiguity problem.

Generative adversarial networks (GAN)~\cite{ian_gan} have been proposed to generate photorealistic images. SR models based on the principles of GAN~\cite{esrganplus, ma_SPSR, details_or_artifacts, srooe_Park_2023_CVPR, Korkmaz_2024_CVPR} are also one-to-one mappings that generate a single SR image (per~$\lambda$).
It is well known that GANs can hallucinate HF details. While some of these hallucinations are readily identifiable as artifacts by human observers, others may appear convincingly realistic despite being fabricated. Consequently, GAN-based SR models fail to provide trustworthy solutions for resolving the ambiguity problem.

\subsection{One-to-Many SR Inference}
Likelihood-based training of SR models that favor accurate density estimation, such as variational autoencoders~\cite{vae_liu2021variational} and normalizing flow methods \cite{jo2021srflowda, fsncsr_song2022fs}, have been introduced to generate a diverse set of SR images from a single LR image. These approaches offer notable benefits compared to GAN-based methods, including stable training and monotonic convergence; however, they produce images with low fidelity scores. Similarly, autoregressive models \cite{arm_pmlr-v119-chen20s} excel in density estimation but suffer from slow inference times due to their sequential sampling processes. In addition, pixel-based image representations require prolonged training times to learn subtle HF details.

Recent advancements in one-to-many SR image generation have been significantly propelled by the development of diffusion models \cite{diff_chung2022improving, SR3_saharia2022image, LDM_rombach2022high, diff_mixture_of_experts_luo2023image}. 
For instance, SR3~\cite{SR3_saharia2022image} achieves remarkable performance through iterative refinement in the pixel domain. Latent diffusion models (LDM) \cite{LDM_rombach2022high} perform diffusion process in the latent space to generate high-resolution SR images, while StableSR \cite{wang2024exploiting} introduces a controllable feature wrapping module that balances quality and fidelity during the inference. PASD \cite{yang2023pasd} enhances stable diffusion by employing feature warping and cross-attention mechanisms to reconstruct high-quality images. SeeSR \cite{lora_diff_realsr_wu2024seesr} improves generative capabilities via semantic prompts, whereas SinSR \cite{wang2024sinsr} accelerates the process by distilling text-to-image models into a single-step SR generation. Despite these advancements, current diffusion models face several challenges, including complex two-stage pipelines, high computational requirements for training, and the emergence of unnatural artifacts that lead to unreliable and ambiguous SR outputs.

Traditional diffusion-SR methods typically involve training models from scratch using LR images as additional inputs \cite{diff_li2022srdiff, LDM_rombach2022high, SR3_saharia2022image, wang2024exploiting}. While effective, this approach is computationally intensive and risks compromising generative priors. Alternative methods \cite{diff_chung2022improving, diff_song2023pseudoinverseguided} circumvent the training process by introducing constraints into the reverse diffusion process of pre-trained synthesis models. However, these methods often struggle with the design of effective constraints due to limited prior knowledge of image degradations, which hinders their generalizability. Therefore, in this work, we employ a pre-trained LDM for $\times$4 SR to avoid lengthy training and propose a framework for reliable sample selection. Our approach addresses the challenge of achieving trustworthy SR by combining the diversity offered by diffusion models with the robustness of ensembling selected output samples.


\subsection{Vision-Language Models (VLM)}
Recently, VLMs~\cite{gpt4_achiam2023gpt, blip2_li2023blip, li2022blip} have gained traction as image interpretation and evaluation tools across various domains, including real-world scene understanding and medical image analysis. These models~\cite{vlm_chen2023iqagpt, vlm_survey_zhang2024vision, vlm_zhang2024quality} excel at automatically describing everyday images by focusing on their semantic content and assessing quality without relying on pixel-based similarity measures. In medical imaging, VLMs have been employed to interpret complex visual data—such as X-rays and MRIs—by generating descriptive captions or answering clinical queries, thereby aiding diagnosis~\cite{vlm_medical_brin2024assessing, vlm_medical_li2024chatgpt}.

Unified VLMs, Bootstrapping Language-Image Pre-training (BLIP) \cite{li2022blip} and its variant allowing querying transformer BLIP-2 \cite{blip2_li2023blip}, excel in both vision-language understanding and generation. They leverage a bootstrapped pre-training framework combining image-text contrastive learning with text generation for effectively handling both image descriptions and question-answering tasks. A more recent advancement, Generative Pre-trained Transformer 4 Omni (GPT-4o) \cite{gpt4_achiam2023gpt}, builds on the transformer architecture and extends its capabilities to multimodal tasks, enabling it to interpret and reason about visual inputs with remarkable accuracy. We posit that the~growing versatility and robustness of VLMs in evaluating images across diverse contexts makes them a suitable choice as an image evaluation tool for reliable information extraction from a diverse set of SR image samples.

\subsection{Relation to our Prior Work and Novelty}
In our previous work \cite{korkmaz2022perception}, we showed ensembling samples generated by a flow model by pixel-wise averaging results in a solution~with a more desirable fidelity vs. perceptual quality trade-off. However, this work neither addresses the sample selection problem nor the~trustworthiness of the solution, which was only evaluated by PSNR vs. LPIPS or PSNR vs. Perception Index~(PI) \cite{blau20182018} plots.
In~\cite{korkmaz2024_icip}, we proposed a human-in-the-loop method for sample selection in the SR space spanned by an LDM to ensemble only selected samples into a trustworthy SR image. However, human feedback for sample selection is costly and time-consuming. In~this work, we introduce a fully automated approach by replacing human feedback with the assessment of SR samples by large VLMs for the selection of reliable samples. This~is~the~first paper to leverage VLMs in selecting reliable diffusion SR samples, offering a novel scalable solution to the trustworthy SR problem. We demonstrate that VLMs can effectively evaluate and interpret diffusion samples through natural language prompts, allowing us to ensemble the most frequently selected consistent SR samples. Additionally, we propose the Trustworthiness Score (TWS)—a hybrid metric that jointly accounts for semantic alignment (CLIP), structural similarity (SSIM), and wavelet-based artifact sensitivity. This enables quantitative validation of VLM selections. In other words, we present a significantly extended and fully automated framework that replaces human selection with VLM-guided evaluation, but more importantly, introduces novel components that extend beyond automation by introducing novel quantitative measures.


\section{Resolving Ambiguity by Sample Selection in the SR Space Generated by Diffusion Models}
\label{method}
The proposed trustworthy SR framework, depicted in Fig.~\ref{fig:arch}, consists of three steps: i) generating a set of SR samples~by diffusion models, ii) reliable sample selection by VLMs, and iii)~ensembling selected samples to generate single trustworthy SR solution.
This section first addresses the generation of an SR sample space by an LDM, followed by a discussion of fully-automated  selection of reliable SR samples. 

\subsection{SR Space Generated by Latent Diffusion Models (LDM)}
\begin{figure}[t!]
\centering
\includegraphics[width=\linewidth]{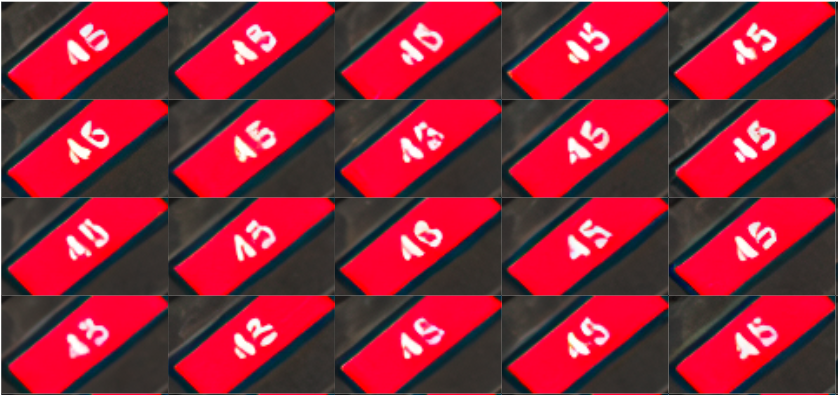}
\caption{Illustration of diversity of SR samples generated by LDM~\cite{LDM_rombach2022high}. Some samples do resemble the ground truth ``45" while others contain a variety of artifacts causing ambiguity.}
\label{fig:45_samples} 
\end{figure}

\begin{figure}[t!]
\centering
\includegraphics[width=0.85\linewidth]{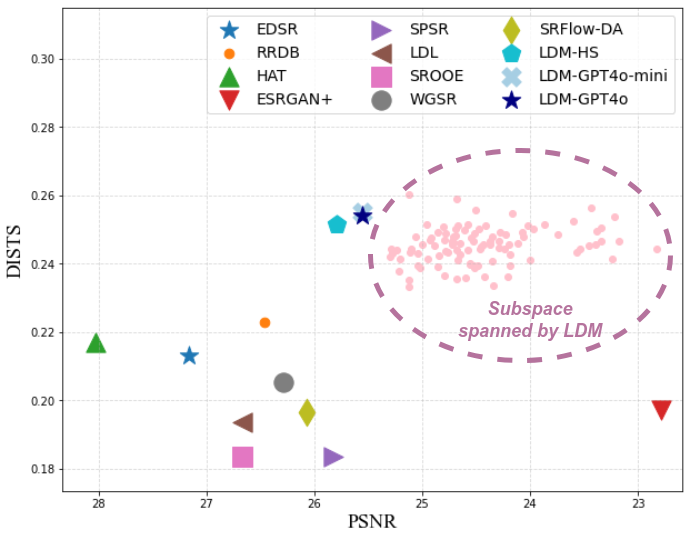} \vspace{-4pt}
\caption{Position of samples generated by LDM~\cite{LDM_rombach2022high}, the proposed ensembled solutions, and other state-of-the-art methods on the DISTS-PSNR plane. We note that perception-distortion tradeoff with respect to popular metrics does not correlate well with trustworthiness and/or visual quality of the solution.}
\label{fig:pd_tradeoff}
\end{figure}

The LDM~\cite{LDM_rombach2022high} performs the diffusion process in a low-dimensional latent space to generate a diverse set of SR samples in a computationally efficient way. We employ the~LDM to generate a set of SR samples, some of which are shown in Fig.~\ref{fig:45_samples}, by using different seeds at the inference time to sample from the distribution learned by a pre-trained model. The distribution of these samples in the DISTS vs. PSNR plane is shown in Fig.~\ref{fig:pd_tradeoff}. It can be seen that diffusion-based SR methods generate a diverse set of SR image samples exhibiting rich texture but the samples may contain hallucinations as depicted in Fig.~\ref{fig:45_samples}. Consequently, selecting a realization from this set at random does not ensure a trustworthy SR solution. Furthermore, none of the widely-used objective evaluation metrics, such as PSNR, LR-Consistency \cite{learning_SR_space_2021}, SSIM, LPIPS \cite{lpips}, and DISTS \cite{dists}, consistently correlate with the reliability of information content in SR images to help trustworthy sample selection. Hence, in the following, we propose leveraging VLMs to enhance the selection of reliable SR samples.

 \vspace{-2pt}

\subsection{Sample Selection via Querying Vision-Language Models}
Rather than relying on human feedback for the selection of reliable diffusion SR samples as presented in our earlier work~\cite{korkmaz2024_icip}, we propose leveraging VLMs ~\cite{vlm_survey_zhang2024vision, li2022blip, blip2_li2023blip, gpt4_achiam2023gpt} including BLIP~\cite{li2022blip} (Bootstrapping Language-Image Pre-training for Unified Vision-Language Understanding and Generation) and GPT4~\cite{gpt4_achiam2023gpt} (Generative Pre-trained Transformer) for the evaluation of SR samples.  Our VLM-based sample selection approach utilizes the capabilities of BLIP-2~\cite{blip2_li2023blip} and GPT-4o~\cite{gpt4_achiam2023gpt} to evaluate and rank SR images generated by LDM. We demonstrate that VLMs provide a practical and robust alternative to human feedback to automate evaluation of SR samples at scale.

We identify two specific tasks for evaluating SR image samples : 1) Sample selection for digit identification, 2) Selection of visually appealing (artifact-free) natural image samples. To facilitate this process, we employ tailored natural language prompts for each task, guiding the VLMs in their assessment and categorization of the SR samples.


For Task 1, which involves evaluating super-resolution (SR) samples for digit identification, we establish the context and utilize prompts such as: ``What is the number in this image?" and ``On a scale of 1 to 100, how certain are you that this number is a clear representation of the digit 5?". When the context is set properly, GPT-4o \cite{gpt4_achiam2023gpt} consistently identifies images that are less prone to artifacts and semantically accurate, completing the digit identification task efficiently. For example, when asked to identify digits, it provided reliable interpretations such as: \textit{``All of these images depict a single character that seems to be a variation of the digit `5' or `6' in a somewhat distorted or stylized font".} Subsequently, the model ranked the images and selected the top-5 and top-1 best samples for further ensembling.

For Task 2, which involves selecting visually appealing samples of natural images,
we use prompts such as: ``Which image contains fewer artifacts and is visually more appealing?" or ``Does this image appear natural-looking to human perception?" By employing clear instructions and providing relevant context through in-context learning, the GPT-4o model demonstrates a notable ability to adapt quickly, delivering structured and reliable outputs. For example, when tasked with analyzing 100 images provided in batches of 10, GPT-4o \cite{gpt4_achiam2023gpt} efficiently responded with detailed explanations, identified artifacts, and ranked the images based on clarity and visual appeal. A sample prompt provided to GPT-4o was: \textit{``I will provide a total of 100 images in 10 batches. For each batch, provide detailed explanations on the appearance of the images. Assess whether they contain artifacts, and determine if they are clear enough to be considered natural-looking by human perception. After reviewing all of the images, select the top-5 and top-1 best-looking images, prioritizing those with fewer artifacts. Also, provide the batch and image number of each selection. Finally, identify the worst-quality image and specify its batch and image number"}. The GPT-4o model responds with highly structured and reliable outputs and after receiving all the images, when asked to select the Top-5 best images, the GPT-4o model consistently delivers a trustworthy solution. 

In both tasks, the selected SR samples are then ensembled by averaging to produce a final SR output that effectively balances fidelity and perceptual quality. To validate the VLM-based approach, we compared the ensembled images against human preferences. Human participants were tasked with the same evaluations as the VLMs, including ranking the samples for clarity, naturalness, and artifact reduction. This direct comparison revealed a strong alignment between human and VLM evaluations, confirming that VLMs serve as a robust and cost-effective alternative to manual feedback in the selection of diffusion samples for SR tasks.

\begin{table}
\caption{VLMs were tasked with identifying the specific digit by querying ``What is the number?" over 324 generated SR samples (MNIST \cite{deng2012mnist} digit 5). Separately, 65 participants were asked to select two samples from the same SR samples that are most helpful to identify the digit as `5' or `6'.}  \vspace{-6pt}
\begin{center}
\begin{adjustbox}{width=0.46\textwidth} 
\begin{tabular}{llll}
\toprule
 & as ``5" & as ``6"  & as others\\
\midrule
BLIP \cite{li2022blip} & 311 (95.9\%) & 10 (3.1\%) &  3 (0.9\%) \\
\midrule
$\#$ of People & 49 (75.4\%) & 16 (24.6\%) & -\\
\bottomrule
\end{tabular}
\end{adjustbox}
\label{table:5or6_results}
\end{center}  
\end{table}

\begin{figure}
\centering
\begin{subfigure}{0.03\textwidth}
    \begin{subfigure}{\textwidth}
        \includegraphics[width=\textwidth]{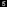} 
    \end{subfigure}
    \caption*{LR \\ input}
\end{subfigure}
\begin{subfigure}{0.075\textwidth}
     \begin{subfigure}{\textwidth}
        \includegraphics[width=\textwidth]{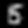} 
    \end{subfigure}
    \caption*{Bicubic Ups.}
\end{subfigure}
\begin{subfigure}{0.075\textwidth}
     \begin{subfigure}{\textwidth}
        \includegraphics[width=\textwidth]{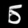} 
    \end{subfigure}
    \caption*{ HAT \\ \cite{hat_chen2023activating}}
\end{subfigure}
\begin{subfigure}{0.075\textwidth}
    \begin{subfigure}{\textwidth}
        \includegraphics[width=\textwidth]{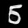} 
    \end{subfigure}
\caption*{ SROOE \\ \cite{srooe_Park_2023_CVPR}}
\end{subfigure}
\begin{subfigure}{0.075\textwidth}
     \begin{subfigure}{\textwidth}
        \includegraphics[width=\textwidth]{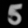} 
    \end{subfigure}
    \caption*{Human Avg. ``5"}
\end{subfigure}
\begin{subfigure}{0.075\textwidth}
    \begin{subfigure}{\textwidth}
        \includegraphics[width=\textwidth]{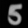} 
    \end{subfigure}
\caption*{VLM Avg.``5"}
\end{subfigure}
\caption{Resolving ambiguity in SR: Identification of the digit from the LR image is ambiguous. Results of the state-of-the-art methods HAT \cite{hat_chen2023activating} (Regressive) and SROOE \cite{srooe_Park_2023_CVPR} (GAN-SR) are also ambiguous. However, the average of five most selected samples by both human participants and VLMs enable mitigating ambiguity yielding a trusthworthy SR solution.}
\label{fig:5or6} 
\end{figure}

\begin{figure*}
\centering
\begin{subfigure}{0.075\textwidth}
     \begin{subfigure}{\textwidth}
        \includegraphics[width=\textwidth]{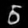} \\ \footnotesize  EDSR \\
        \cite{EDSR2017} \\ 20.28 \\ 0.146 \\ 0.2188
    \end{subfigure} 
     \begin{subfigure}{\textwidth}
        \includegraphics[width=\textwidth]{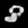} \\ \footnotesize  EDSR \\ \cite{EDSR2017} \\ 18.74 \\ 0.136 \\ 0.1502
    \end{subfigure}     
    \begin{subfigure}{\textwidth}
        \includegraphics[width=\textwidth]{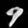} \\ \footnotesize  EDSR \\ \cite{EDSR2017} \\ 19.89 \\ 0.157 \\ 0.3304
    \end{subfigure}
\end{subfigure}
\begin{subfigure}{0.075\textwidth}
    \begin{subfigure}{\textwidth}
        \includegraphics[width=\textwidth]{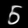} \\ \footnotesize  RRDB \\ \cite{wang2018esrgan} \\ 20.99 \\ 0.102 \\ 0.1810
    \end{subfigure}
    \begin{subfigure}{\textwidth}
        \includegraphics[width=\textwidth]{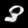} \\ \footnotesize  RRDB \\ \cite{wang2018esrgan} \\ 18.43 \\ 0.146 \\ 0.1898
    \end{subfigure}
    \begin{subfigure}{\textwidth}
        \includegraphics[width=\textwidth]{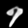} \\ \footnotesize  RRDB \\ \cite{wang2018esrgan} \\ 19.32 \\ 0.153 \\ 0.3465
    \end{subfigure}
\end{subfigure}
\begin{subfigure}{0.075\textwidth}
    \begin{subfigure}{\textwidth}
        \includegraphics[width=\textwidth]{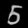} \\ \footnotesize  HAT \\ \cite{hat_chen2023activating} \\ 20.46 \\ 0.125 \\ 0.2155
    \end{subfigure}
    \begin{subfigure}{\textwidth}
        \includegraphics[width=\textwidth]{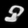} \\\footnotesize  HAT \\ \cite{hat_chen2023activating} \\ 17.96 \\ 0.136 \\ 0.1866
    \end{subfigure}
    \begin{subfigure}{\textwidth}
        \includegraphics[width=\textwidth]{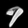} \\\footnotesize  HAT \\ \cite{hat_chen2023activating} \\ 17.82 \\ 0.191 \\ 0.2834
    \end{subfigure}
\end{subfigure}
\begin{subfigure}{0.075\textwidth}
    \begin{subfigure}{\textwidth}
        \includegraphics[width=\textwidth]{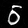} \\ \footnotesize  ESRGAN+ \cite{esrganplus} \\ 19.00 \\ 0.108 \\ 0.0680
    \end{subfigure}
    \begin{subfigure}{\textwidth}
        \includegraphics[width=\textwidth]{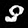} \\ \footnotesize  ESRGAN+ \cite{esrganplus} \\ 16.61 \\ 0.122 \\  0.1142
    \end{subfigure}
    \begin{subfigure}{\textwidth}
        \includegraphics[width=\textwidth]{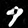} \\ \footnotesize  ESRGAN+ \cite{esrganplus} \\ 17.51 \\ 0.152 \\ 0.1540
    \end{subfigure}
\end{subfigure}
\begin{subfigure}{0.075\textwidth}
    \begin{subfigure}{\textwidth}
        \includegraphics[width=\textwidth]{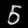} \\ \footnotesize  LDL \\ \cite{details_or_artifacts} \\21.29 \\ 0.095 \\ 0.1731
    \end{subfigure}
    \begin{subfigure}{\textwidth}
        \includegraphics[width=\textwidth]{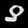} \\ \footnotesize  LDL \\ \cite{details_or_artifacts} \\17.84 \\ 0.118 \\ 0.1680
    \end{subfigure}
    \begin{subfigure}{\textwidth}
        \includegraphics[width=\textwidth]{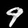} \\ \footnotesize  LDL \\ \cite{details_or_artifacts} \\ 22.65 \\ 0.087 \\ 0.2784
    \end{subfigure}
\end{subfigure} 
\begin{subfigure}{0.075\textwidth}
    \begin{subfigure}{\textwidth}
        \includegraphics[width=\textwidth]{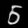} \\ \footnotesize  SROOE \\ \cite{srooe_Park_2023_CVPR} \\ 20.93 \\ 0.106 \\ 0.1838
    \end{subfigure}
    \begin{subfigure}{\textwidth}
        \includegraphics[width=\textwidth]{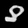} \\ \footnotesize  SROOE \\ \cite{srooe_Park_2023_CVPR} \\ 18.37 \\ 0.125 \\ 0.1753
    \end{subfigure}
    \begin{subfigure}{\textwidth}
        \includegraphics[width=\textwidth]{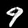} \\ \footnotesize  SROOE \\ \cite{srooe_Park_2023_CVPR} \\ 22.53 \\ 0.078 \\ 0.2446
    \end{subfigure}
\end{subfigure}
\begin{subfigure}{0.075\textwidth}
    \begin{subfigure}{\textwidth}
        \includegraphics[width=\textwidth]{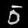} \\ \footnotesize  HCFlow++ \cite{hcflow_liang21hierarchical} \\ 19.45 \\ 0.121 \\  0.1604
    \end{subfigure}
    \begin{subfigure}{\textwidth}
        \includegraphics[width=\textwidth]{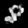} \\ \footnotesize  HCFlow++ \cite{hcflow_liang21hierarchical} \\ 17.85 \\ 0.204 \\  0.1419
    \end{subfigure}
    \begin{subfigure}{\textwidth}
        \includegraphics[width=\textwidth]{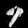} \\ \footnotesize  HCFlow++ \cite{hcflow_liang21hierarchical} \\ 17.55 \\ 0.208 \\ 0.1369
    \end{subfigure}
\end{subfigure}
\begin{subfigure}{0.075\textwidth}
    \begin{subfigure}{\textwidth}
        \includegraphics[width=\textwidth]{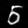} \\ \footnotesize  SRFlowDA \cite{jo2021srflowda} \\ 21.33 \\ 0.099 \\ 0.1751
    \end{subfigure}
    \begin{subfigure}{\textwidth}
        \includegraphics[width=\textwidth]{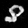} \\ \footnotesize  SRFlowDA \cite{jo2021srflowda} \\ 17.68 \\ 0.151 \\ 0.2204
    \end{subfigure}
    \begin{subfigure}{\textwidth}
        \includegraphics[width=\textwidth]{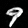} \\ \footnotesize  SRFlowDA \cite{jo2021srflowda} \\ 20.88 \\ 0.118 \\ 0.2879
    \end{subfigure}
\end{subfigure}
\begin{subfigure}{0.075\textwidth}
    \begin{subfigure}{\textwidth}
        \includegraphics[width=\textwidth]{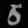} \\ \footnotesize LDM \\ \cite{LDM_rombach2022high} \\ 16.89 \\ 0.212 \\ 0.1987
    \end{subfigure}
    \begin{subfigure}{\textwidth}
        \includegraphics[width=\textwidth]{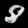} \\ \footnotesize LDM \\ \cite{LDM_rombach2022high} \\ 16.73 \\ 0.190 \\  0.1239
    \end{subfigure}
    \begin{subfigure}{\textwidth}
        \includegraphics[width=\textwidth]{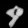} \\ \footnotesize LDM \\ \cite{LDM_rombach2022high} \\ 16.24 \\ 0.239 \\  0.3564
    \end{subfigure}
\end{subfigure}
\begin{subfigure}{0.075\textwidth}
    \begin{subfigure}{\textwidth}
        \includegraphics[width=\textwidth]{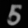} \\ \footnotesize LDM-HS \\ \cite{korkmaz2024_icip} \\ 17.62 \\ 0.215 \\ 0.2177
    \end{subfigure}
    \begin{subfigure}{\textwidth}
        \includegraphics[width=\textwidth]{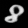} \\ \footnotesize LDM-HS \\ \cite{korkmaz2024_icip} \\ 17.07 \\ 0.151 \\ 0.1847
    \end{subfigure}
    \begin{subfigure}{\textwidth}
        \includegraphics[width=\textwidth]{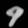} \\ \footnotesize LDM-HS \\ \cite{korkmaz2024_icip} \\ 17.63 \\ 0.264 \\ 0.3878
    \end{subfigure}
\end{subfigure}
\begin{subfigure}{0.075\textwidth}
    \begin{subfigure}{\textwidth}
        \includegraphics[width=\textwidth]{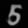} \\ \footnotesize LDM-VLM \\ (Ours) \\ 17.83 \\ 0.264 \\ 0.2206
    \end{subfigure}
    \begin{subfigure}{\textwidth}
        \includegraphics[width=\textwidth]{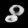} \\ \footnotesize LDM-VLM \\ (Ours) \\ 16.72 \\ 0.231 \\ 0.1793
    \end{subfigure}
    \begin{subfigure}{\textwidth}
        \includegraphics[width=\textwidth]{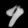} \\ \footnotesize LDM-VLM \\ (Ours) \\ 17.15 \\ 0.262 \\ 0.3557
    \end{subfigure}
\end{subfigure}
\begin{subfigure}{0.075\textwidth}
    \begin{subfigure}{\textwidth}
        \includegraphics[width=\textwidth]{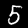} \\ \footnotesize HR \\ (5) \\ 
         PSNR$\uparrow$ \\ DISTS$\downarrow$\cite{dists} \\ TWS$\uparrow$
     \end{subfigure}
    \begin{subfigure}{\textwidth}
        \includegraphics[width=\textwidth]{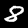} \\ \footnotesize HR \\ (8) \\ PSNR$\uparrow$ \\ DISTS$\downarrow$\cite{dists} \\ TWS$\uparrow$
    \end{subfigure}
    \begin{subfigure}{\textwidth}
        \includegraphics[width=\textwidth]{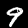} \\ \footnotesize HR \\ (9) \\ PSNR$\uparrow$ \\ DISTS$\downarrow$\cite{dists} \\ TWS$\uparrow$
    \end{subfigure}
\end{subfigure} 
\caption{Visual comparison of the proposed LDM-VLM method vs. the state-of-the-art regressive, GAN-based, flow-based, and diffusion-based SR methods on MNIST dataset \cite{deng2012mnist}. Observe that our proposed methods provide reliable SR images, but popular quantitative metrics cannot capture the nuances of visual artifacts or trustworthiness.}
\label{fig:mnist} 
\end{figure*}

\begin{table*}[t]
\caption{The number of times the popular Vision-Language Models identified the image as ``number" when prompted with the query ``What is the number?" for each of 100 SR samples generated by the LDM (some of which are depicted in Fig.~\ref{fig:45_samples})}  \vspace{-5pt}
\begin{center}
\begin{adjustbox}{width=\textwidth}
\begin{tabular}{cccccccccccccc}
\toprule
 & as``45" & as ``46" & as ``40" & as ``41"& as ``42"& as ``43"& as ``44"& as ``47" & as ``48" & as ``49" & as ``4" & as ``5" & others \\
\midrule
BLIP \cite{li2022blip} & 5 & 1 & 1 & - & 9 & - & 3 & - & - & 1 & 33 & 21 & 26 \\
BLIP-2 \cite{blip2_li2023blip}& 16 & 2 & - & - & 1 & - & 2 & - & - & - & 36 & 8 & 35 \\
Llama OCR \cite{llama_ocr} & 9 & 7 & - & - & - & - & -& - & 3 & - & - & - & 81 \\
GPT4o-mini \cite{gpt4_achiam2023gpt} & 39 & 15 & 9 & 3 & 4 & 12 & 1 & 1 & 10 & 6 & - & -& -\\
GPT4o \cite{gpt4_achiam2023gpt} & 44 & 17 & 5 & 4 & 3 & 17 & 2 & - & 6 & 2  & - & - & -\\
$\#$ of People & 33 (50.8\%) & 7 (10.8\%) & 3 (4.6\%) & - & - & 16 (24.6\%) & - & 1 (1.5\%) & - & 1 (1.5\%)  & - &  & 4 (6.2\%)\\
\bottomrule
\end{tabular}
\end{adjustbox}
\label{table:vlm_45_results}
\end{center}  
\end{table*}

\subsection{Trustworthiness Score for SR Evaluation}
Evaluating the reliability of SR images remains a challenging problem, particularly in information-critical applications where perceptual fidelity does not necessarily imply correctness. Conventional full-reference image quality metrics such as PSNR, SSIM, and perceptual measures like LPIPS and DISTS often fail to provide a meaningful assessment of trustworthiness, as they either lack correlation with human perception in semantic tasks or become impractical when the ground truth is unavailable. To address this issue, we introduce a hybrid similarity metric that quantifies the trustworthiness of an SR image by evaluating three key aspects: semantic consistency, structural integrity, and artifact suppression. Our metric integrates CLIP-based similarity to ensure semantic correctness, edge-based SSIM to preserve structural details, and a wavelet-based score to penalize high-frequency artifacts and blurring effects.

Given an input high-resolution image \( I_{HR} \) and a super-resolved image \( I_{SR} \), the trustworthiness metric \( TWS(I_{HR}, I_{SR}) \) is formulated as follows:

\[
T(I_{HR}, I_{SR}) = \lambda_{CLIP} S_{CLIP} + \lambda_{edge} S_{edge} - \lambda_{wavelet} S_{wavelet},
\]

\noindent
where \( S_{CLIP} \) represents the semantic similarity computed using CLIP embeddings, \( S_{edge} \) denotes the structural similarity derived from edge-based SSIM, and \( S_{wavelet} \) is the normalized wavelet-based artifact score.

The semantic similarity score \( S_{CLIP} \) is computed by extracting feature embeddings from CLIP and evaluating their cosine similarity:


\[  
S_{CLIP} = \frac{E_{CLIP}(I_{HR}) \cdot E_{CLIP}(I_{SR})}{\|E_{CLIP}(I_{HR})\|\|E_{CLIP}(I_{SR})\|}
\]

\noindent
where \( E_{CLIP}(I) \) denotes the feature representation obtained from the CLIP model. This measure ensures that the SR image maintains the high-level semantic meaning of the original.

The structural similarity \( S_{edge} \) is computed using edge-based SSIM. This approach focuses on preserving fine-grained details such as edges and contours, which are critical for legibility and feature preservation in information-dense images. Edge maps \( E(I) \) are extracted using an Sobel edge-detection operator and SSIM is then applied:

\[
S_{edge} = SSIM(E(I_{HR}), E(I_{SR})).
\]

To quantify the presence of artifacts and blurring, we introduce a wavelet-based artifact score. This is computed by performing a multi-level discrete wavelet decomposition on the grayscale version of \( I_{SR} \) using the Daubechies-19 (`db19') wavelet. The decomposition separates the image into low and high-frequency components at multiple levels. The high-frequency sub-bands, which capture fine details, are aggregated to compute the total high-frequency energy. The sum of absolute values of these coefficients is then normalized by the total number of pixels in the image:

\[
S_{wavelet} = \sum_{j=1}^{L} \sum_{c} ||W_c^{(j)}(I_{SR})||_1,
\]

\noindent
where \( L = 2\) is the number of decomposition levels and \( W_c^{(j)}(I_{SR}) \) represents the wavelet coefficients at level \( j \) for high-frequency sub-band \( c \). Higher values of \( S_{wavelet} \) indicate a stronger presence of high-frequency distortions, such as noise, ringing artifacts, or unnatural edges. The negative weighting of \( S_{wavelet} \) ensures that increased artifacts lead to a lower trustworthiness score.

Since the TWS is a hybrid metric composed of multiple components with different scales, we normalize each component to the [0, 1] range to ensure comparability. The relative importance of each component was determined through a targeted weight search on a single representative case—image 45 in Figure~\ref{fig:first_img}. The resulting weights, $\lambda_{\text{CLIP}} = 0.2$, $\lambda_{\text{edge}} = 0.3$, and $\lambda_{\text{wavelet}} = 0.5$, were then fixed and applied uniformly across all subsequent evaluations.

To evaluate the generalizability of this formulation, we applied TWS to a diverse set of domains, including digit recognition (MNIST~\cite{deng2012mnist}), character-level restoration, and natural image super-resolution using datasets such as Set14~\cite{set14_cite}, BSD100~\cite{bsd100_cite}, and DIV2K~\cite{Agustsson_2017_CVPR_Workshops}. Across all settings, TWS consistently favored perceptually accurate and semantically faithful outputs, demonstrating strong alignment with human judgments even in the absence of ground-truth references.

Overall, the proposed metric provides a robust and scalable measure of trustworthiness in selecting  diffusion SR samples. By jointly capturing semantic fidelity, structural consistency, and artifact suppression, TWS enables automated selection of reliable SR outputs in real-world scenarios,--particularly for information-critical applications--without requiring human-labeled ground truth. 

\section{Experiments}
\label{exp}
\subsection{Experimental Setup}
We selected widely used datasets as benchmarks for our study: MNIST~\cite{deng2012mnist}, BSD100 \cite{bsd100_cite}, Urban100 \cite{urban100_cite} and DIV2K~\cite{Agustsson_2017_CVPR_Workshops}. For the MNIST dataset, where the original images are 28$\times$28 grayscale, we downsampled the images by a factor of 4 in each dimension using Matlab's bicubic kernel, resulting in 7$\times$7 LR images. Since the LDM~\cite{LDM_rombach2022high} was pre-trained to super-resolve images from 128$\times$128 to 512$\times$512, we adapted the 7$\times$7 MNIST images by repeating each sample 18 times horizontally and vertically (replicating the last two rows and columns as needed) to form 128$\times$128 LR images. These processed images, each containing a grid of 18$\times$18 MNIST digit samples (384 digits in total), were then fed into the LDM. As the LDM performs a one-to-many mapping, the resulting SR images showcase a variety of upsampled digits for each input. For BSD100, Urban100 and DIV2K, 128$\times$128 RGB LR patches were cropped from the original LR images and directly fed into the pre-trained LDM to generate 512$\times$512 SR samples. To ensure diversity in the generated outputs, this process was repeated 100 times for each natural image.

In summary, the SR space generated by the LDM model comprises 324 samples for MNIST digits (spanning multiple variations of digit representations) and 100 diverse SR samples for each natural image from the BSD100, Urban100 and DIV2K dataset. This extensive collection provides a robust foundation for evaluating our sample selection approach.

\subsection{Automated Assessment of Samples by VLM}
\subsubsection{Results on Digits}
We evaluated BLIP \cite{li2022blip}, BLIP-2 \cite{blip2_li2023blip}, Llama OCR \cite{llama_ocr} and two variants of the GPT-4 model \cite{gpt4_achiam2023gpt}, GPT-4o and GPT-4o-mini, to assess their ability to identify digits in SR tasks by querying ``What is the number?". For the MNIST dataset, we provided a set of 324 diffusion-generated SR samples to the VLMs for digit identification. The results of this evaluation are presented in Table \ref{table:5or6_results}. The BLIP model \cite{li2022blip} demonstrated a high level of consistency, accurately identifying the digit ``5" in 95.9$\%$ of the responses when tested on the diffusion samples. To refine the selection process further, we employed a chain-of-thought approach by asking the model, ``On a scale of 1 to 100, how certain are you that this number is a 5?" This method allowed us to reduce the number of images to 28 based on confidence levels. We then input these 28 samples into GPT-4o \cite{gpt4_achiam2023gpt}, instructing it to select the top 5 natural-looking images with fewer artifacts. The averaged result of this selection is depicted in Fig. \ref{fig:5or6}. Additionally, we applied a combined BLIP \cite{li2022blip} + GPT-4o \cite{gpt4_achiam2023gpt} method for selecting MNIST samples of other digits, with results shown in Fig.~\ref{fig:mnist}. We integrated these two VLMs because while BLIP~\cite{li2022blip} alone struggled to identify the top 5 images, providing all 324 images to GPT-4o \cite{gpt4_achiam2023gpt} would have been computationally expensive. Therefore, BLIP \cite{li2022blip} was used to pre-filter the set, and GPT-4o \cite{gpt4_achiam2023gpt} finalized the selection.

\begin{table*}[t!]
\begin{center}
\begin{adjustbox}{width=0.95\textwidth}
\begin{tabular}{llccccccccc}
\toprule
 & SR Model & PSNR$\uparrow$ & LR Consistency $\uparrow$ & SSIM$\uparrow$ & LPIPS$\downarrow$ & LPIPS$_{\text{VGG}}$$\downarrow$ & PieAPP$\downarrow$ & DISTS$\downarrow$ & NRQM$\uparrow$ & TWS$\uparrow$ \\
\midrule
 \multirow{3}{*}{Regressive} & EDSR &  25.962 & 43.047 & 0.803 &  0.115 &  0.231 &  0.901 &  0.194 &  5.142 &  0.3542\\
 Regressive &  RRDB &  25.316 & 39.508 & 0.788 &  0.103 &  0.225 &  0.799 &  0.187 &  5.850 & 0.3336\\
 &  HAT &  27.408 & 44.673 & 0.826 &  0.089 &  0.201 &  0.686 &  0.179 &  5.518 & 0.3480\\ \hline
 \multirow{5}{*}{GAN-based}&  ESRGAN+ &  22.666 &  31.718 & 0.716 &  0.083 &  0.224 &  0.292 &  0.168 &  7.757 &  0.2735 \\
 &  SPSR &  24.760 & 36.520 & 0.762 &  0.063 &  0.184 &  0.523 &  0.138 &  7.159 &  0.3158 \\ 
  &  LDL &  27.194 & 43.360 & 0.852 &  0.053 &  0.145 &  0.396 &  0.125 &  7.079 & 0.3505\\ 
&  SROOE &  25.894 & 41.040 & 0.790 &  0.061 &  0.166 &  0.562 &  0.132 &  6.741 & 0.3348\\ 
&  WGSR & 26.177 & 42.960 & 0.821 & 0.077 & 0.199 & 0.508 & 0.158 & 7.351 & 0.3243\\\hline
 \multirow{2}{*}{Flow-based} &  SRFlowDA &  27.510 & 46.929 & 0.852 &  0.062 &  0.172 &  0.686 &  0.145 &  6.699 & {0.3551}\\ 
&  HCFlow &  25.062 & 43.302 & 0.777 &  0.067 &  0.183 &  0.641 &  0.141 &  6.896 & 0.3257\\ \hline
\multirow{9}{*}{Diffusion-based} & SR3 (random sample) & 21.596 & 25.587 & 0.683 & 0.231 & 0.299 & 2.065 & 0.357 & 6.649 & 0.1033 \\
 & LDM (random sample) &  24.234 & 29.655 & 0.780 &  0.122 &  0.244 &  0.898 &  0.185 &  5.794 & 0.3291\\ 
 & IDM (random sample) &  24.573 & 29.526 & 0.716 & 0.149 & 0.294 & 0.651 & 0.227& 6.496 & 0.2709\\
 & SinSR (random sample) & 23.097 & 28.295 & 0.721 & 0.135 & 0.289 & 1.049 & 0.206 & 6.099 & 0.3000 \\
 & PASD (random sample) & 23.206 & 28.828 & 0.711 & 0.143 & 0.284 & 1.282 & 0.205 & 6.190 & 0.3116 \\
& LDM-HS (ensemble) &  26.047 & 31.447 & 0.823 &  0.141 &  0.227 &  1.120 &  0.194 &  5.195 & 0.3602\\
& LDM-VLM-Top1 & 24.762 & 30.478 & 0.797 & 0.123 & 0.231 & 1.039 & 0.179 & 6.054 & 0.3348\\
& LDM-VLM (ensemble) & 25.927  & 31.561 & 0.824 & 0.139 & 0.228 & 1.123 & 0.194 & 5.181 & 0.3613 \\
\bottomrule
\end{tabular}
\end{adjustbox}
\caption{Performance comparison of different $\times$4 SR methods on 128$\times$128 SR patches from the DIV2K validation~set. The fully automated LDM-VLM provides a practical solution that maintains accuracy with comparable fidelity results. However, observe that there exists a notable divergence between visual accuracy and quantitative measures.}
\label{table:fusion_results}
\end{center}  
\end{table*}

To validate our VLM-based automated evaluation pipeline, we conducted a human evaluation using a mirrored task similar to those employed for assessing the VLM methods.  We invited 65 participants to identify a specific digit from images in the MNIST \cite{deng2012mnist} dataset and to select the two most ``natural" looking numbers from a pool of 324 generated SR samples. Specifically, our human evaluation involved 65 participants comprising a mix of students, researchers, and professionals, including 10 with formal backgrounds in computer vision or image quality assessment. This distribution ensures a balanced evaluation reflecting both expert insight and general perceptual judgment. This task not only evaluates  the perceptual quality but also assesses the SR model's ability to preserve critical information conveyed in the image, specifically the digit itself. As illustrated in Fig. \ref{fig:mnist} the top-5 most selected samples, human feedback confirmed that VLMs can effectively identify trustworthy SR samples, distinguishing ambiguous digits. For instance, while a state-of-the-art method mistakenly produced a ``6" instead of the correct ``5," our approach utilizing LDM-HumanSelection (LDM-HS) produced the accurate digit. Similarly, the digit ``8" that appeared unclear in other methods was correctly identified as an ``8" by our method. 
This approach underscores the limitations of traditional objective metrics and highlights that samples selected by VLMs are as reliable as those selected by humans in information-centric SR applications.

We evaluated a set of 100 diffusion-generated SR samples containing the number ``45" from the Urban100 \cite{urban100_cite} dataset, tasking the VLMs with the query, “What is the number?” The models analyzed SR images and provided their predictions of the most likely digit based on the visual content of each SR image. We recorded the models' responses and assessed the consistency of their predictions. The results for ``45"  diffusion samples from the Urban100 dataset \cite{urban100_cite} are presented in Table \ref{table:vlm_45_results}. Notably, the GPT-4o model outperformed the other models, correctly identifying ``45" in 44\% of the samples, significantly surpassing the accuracy of both BLIP \cite{li2022blip} and BLIP-2 \cite{blip2_li2023blip}. In our further analysis, we observed that while BLIP and its successor BLIP-2 are primarily designed for visual understanding tasks, GPT-4o \cite{gpt4_achiam2023gpt} has been optimized for a broader range of applications, including zero-shot and few-shot learning. This versatility enables GPT-4o to generalize more effectively  from its training data, making it particularly adept at handling ambiguities in SR images where subtle visual cues are critical for accurate digit identification. 

Similarly, we asked 65 participants to identify a specific number from images in the Urban100 \cite{urban100_cite} dataset. A majority of the participants $50.8\%$ answered the number as ``45", aligning with the results obtained from the GPT-4o model, as shown in Table \ref{table:vlm_45_results}. This approach allowed us to assess how effectively the VLMs handle ambiguous or degraded SR outputs, ensuring that the identified numbers correspond with human understanding while minimizing subjective biases. Additionally, participants were tasked with selecting the two most similar samples that represented their answers to acquire a single trustworthy image from the diffusion samples. The average of the top-5 selected samples is shown in Fig.~\ref{fig:first_img}. The results from human evaluations closely matched the selections made by the VLMs, validating that these models are effective in identifying high-quality SR images with accurate information content. 
While human-selected ensembles (VLM-HS) remain effective in capturing perceptual quality, the VLM-ensembled approaches offer several distinct advantages. The selections made by GPT-4o not only demonstrated a high degree of consistency but also provided a scalable and cost-effective alternative to subjective human feedback. This capability is particularly beneficial for large-scale tasks where manual evaluations are impractical or prohibitively expensive. To further validate the quality and reliability of the selected samples, we computed TWS across all evaluated methods. Both LDM-HS and LDM-VLM ensembles achieved significantly higher TWS values than other SR baselines, indicating superior preservation of semantic accuracy, structural consistency, and reduced artifacts. Notably, LDM-VLM achieved the highest average TWS, surpassing even human-guided selection. This underscores the practical feasibility and effectiveness of our automated pipeline in selecting trustworthy SR samples from the diffusion space. Moreover, the structured and repeatable nature of VLM-based evaluation supports its robustness in handling complex visual ambiguities, including those involving subtle distortions or fine details. Overall, these results emphasize the effectiveness of using vision-language models (LDM-VLM) in strategically selecting diffusion model samples to mitigate visual artifacts and enhance overall perceptual quality. Additionally, VLMs can serve as a reliable alternative to human evaluations.

\subsubsection{Results on Natural Images}

\noindent
Table \ref{table:fusion_results} presents quantitative comparison of 4$\times$ SR methods, including our proposed ensembling method using VLM GPT-4o \cite{gpt4_achiam2023gpt} on the DIV2K \cite{Agustsson_2017_CVPR_Workshops} validation set. The model was prompted with the question: ``How visually appealing is the image, considering the presence of fewer artifacts and overall clarity from a human perception standpoint?" We instructed GPT-4o to select the top five (LDM-VLM Top-5) and top one (LDM-VLM Top-1) images that best aligned with this query. The results were then compared to state-of-the-art methods, such as EDSR \cite{EDSR2017}, RRDB \cite{wang2018esrgan}, HAT \cite{hat_chen2023activating}, ESRGAN+ \cite{esrganplus}, SPSR \cite{ma_SPSR}, LDL \cite{details_or_artifacts}, and SROOE \cite{srooe_Park_2023_CVPR}, as well as stochastic SR methods like HCFLow++ \cite{hcflow_liang21hierarchical} and SRFlow-DA \cite{jo2021srflowda}. 

Compared to regressive methods such as EDSR \cite{EDSR2017} and RRDB \cite{wang2018esrgan}, our VLM-based sample selection approach (Top-1 and ensemble) provides on par fidelity scores with significantly improved perceptual quality. Similarly, in terms of perceptual metrics, such as LPIPS \cite{lpips} and DISTS \cite{dists}, LDM-VLM (ensemble) performs on par with or surpasses both GAN-based and flow-based methods. When comparing diffusion-based approaches, LDM-VLM (ensemble) notably outperforms random sampling of outputs from LDM \cite{LDM_rombach2022high} and IDM \cite{implicit_gao2023implicit}. For example, LDM-VLM (ensemble) improves upon LDM (random sample) by +1.6 dB in PSNR and 5.6$\%$ in SSIM, indicating a greater ability to retain fidelity. These improvements underscore the effectiveness of VLM-driven sample selection for maximizing information retention and perceptual quality simultaneously.

While we present quantitative comparison results for our proposed approach, the effectiveness of evaluating visual artifacts in SR tasks cannot rely solely on metrics such as PSNR or other quantitative perceptual scores. Although these metrics provide numerical insights into image quality, they may fail to capture the subtle nuances of visual artifacts effectively. Our results, on the other hand, show that TWS provides a more meaningful indicator of SR performance, with both LDM-HS and LDM-VLM outperforming other methods. Notably, LDM-VLM achieves the highest TWS overall, confirming its effectiveness as a scalable and reliable solution for generating trustworthy SR outputs. As a result, by combining human-like reasoning from VLMs with diffusion models, we achieve a comprehensive framework for trustworthy SR.

\begin{figure*}
\centering
\begin{subfigure}{0.16\textwidth}
     \begin{subfigure}{\textwidth}
        \includegraphics[width=\textwidth]{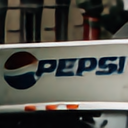} \\ \small  EDSR \cite{EDSR2017}\\ (23.88 / 0.213) \\ 0.3391
    \end{subfigure}
    \begin{subfigure}{\textwidth}
        \includegraphics[width=\textwidth]{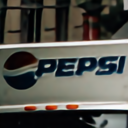} \\ \small  RRDB \cite{wang2018esrgan} \\ (23.81 / 0.214) \\ 0.3269
    \end{subfigure}
     \begin{subfigure}{\textwidth}
            \includegraphics[width=\textwidth]{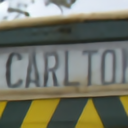} \\ \small  EDSR \cite{EDSR2017} \\ (31.88 / 0.137) \\ 0.3856
    \end{subfigure}
    \begin{subfigure}{\textwidth}
        \includegraphics[width=\textwidth]{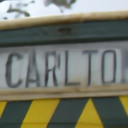} \\ \small  RRDB \cite{wang2018esrgan} \\ (30.51 / 0.121) \\ 0.3770
    \end{subfigure}
\end{subfigure}
\begin{subfigure}{0.16\textwidth}
    \begin{subfigure}{\textwidth}
        \includegraphics[width=\textwidth]{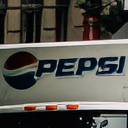} \\ \small  LDL \cite{details_or_artifacts} \\ (23.84 / 0.134) \\ 0.3113
    \end{subfigure}
    \begin{subfigure}{\textwidth}
        \includegraphics[width=\textwidth]{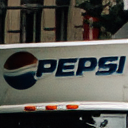} \\ \small  SROOE \cite{srooe_Park_2023_CVPR} \\ (24.60 / 0.122) \\ 0.3331
    \end{subfigure}
    \begin{subfigure}{\textwidth}
        \includegraphics[width=\textwidth]{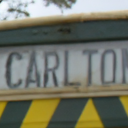} \\ \small  LDL \cite{details_or_artifacts} \\ (32.39 / 0.100) \\ 0.3872
    \end{subfigure}
    \begin{subfigure}{\textwidth}
        \includegraphics[width=\textwidth]{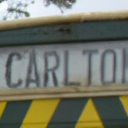} \\ \small  SROOE \cite{srooe_Park_2023_CVPR} \\ (33.06 / 0.085) \\ 0.3834
    \end{subfigure}
\end{subfigure}
\begin{subfigure}{0.16\textwidth}
    \begin{subfigure}{\textwidth}
        \includegraphics[width=\textwidth]{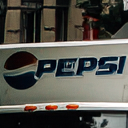} \\\small   HCFlow++ \cite{hcflow_liang21hierarchical} \\ (23.40 / 0.141) \\ 0.3013
    \end{subfigure}
    \begin{subfigure}{\textwidth}
        \includegraphics[width=\textwidth]{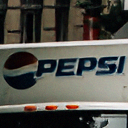} \\\small  SRFlow-DA \cite{jo2021srflowda} \\ (24.81 / 0.143) \\ 0.3252
    \end{subfigure}
    \begin{subfigure}{\textwidth}
        \includegraphics[width=\textwidth]{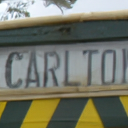} \\\small  HCFlow++ \cite{hcflow_liang21hierarchical} \\ (30.65 / 0.116) \\ 0.4021
    \end{subfigure}
    \begin{subfigure}{\textwidth}
        \includegraphics[width=\textwidth]{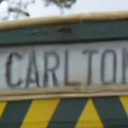} \\\small  SRFlow-DA \cite{jo2021srflowda} \\ (32.14 / 0.116) \\ 0.3878
    \end{subfigure}
\end{subfigure}
\begin{subfigure}{0.16\textwidth}
    \begin{subfigure}{\textwidth}
        \includegraphics[width=\textwidth]{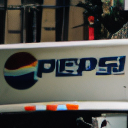} \\ \small  LDM \cite{LDM_rombach2022high} (random) \\ (22.11 / 0.179) \\ 0.2961
    \end{subfigure}
    \begin{subfigure}{\textwidth}
        \includegraphics[width=\textwidth]{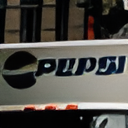} \\ \small  IDM \cite{implicit_gao2023implicit} \\ (21.13 / 0.212) \\ 0.2656
    \end{subfigure}
    \begin{subfigure}{\textwidth}
        \includegraphics[width=\textwidth]{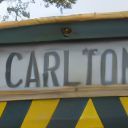} \\ \small LDM \cite{LDM_rombach2022high} (random)\\ (28.49 / 0.169) \\ 0.3774
    \end{subfigure}
    \begin{subfigure}{\textwidth}
        \includegraphics[width=\textwidth]{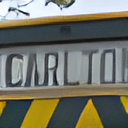} \\ \small IDM \cite{implicit_gao2023implicit} \\ (24.04 / 0.205) \\ 0.3036
    \end{subfigure}
\end{subfigure}
\begin{subfigure}{0.16\textwidth}
    \begin{subfigure}{\textwidth}
        \includegraphics[width=\textwidth]{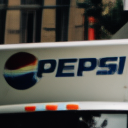} \\ \small LDM-HS \cite{korkmaz2024_icip} \\(23.98 / 0.242) \\ 0.3545
    \end{subfigure}
    \begin{subfigure}{\textwidth}
        \includegraphics[width=\textwidth]{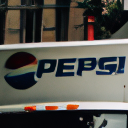} \\ \small LDM-VLM-Top1 \\ (22.90 / 0.289) \\ 0.3177
    \end{subfigure}
    \begin{subfigure}{\textwidth}
        \includegraphics[width=\textwidth]{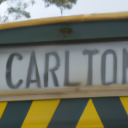} \\ \small LDM-HS \cite{korkmaz2024_icip} \\(26.71 / 0.209) \\ 0.3956
    \end{subfigure}
    \begin{subfigure}{\textwidth}
        \includegraphics[width=\textwidth]{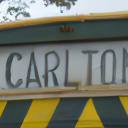} \\ \small LDM-VLM-Top1 \\ (28.19 / 0.297) \\ 0.3798
    \end{subfigure}
\end{subfigure}
\begin{subfigure}{0.16\textwidth}
    \begin{subfigure}{\textwidth}
        \includegraphics[width=\textwidth]{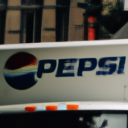} \\ \small LDM-VLM (Ours) \\(23.91 / 0.264) \\ 0.3571
    \end{subfigure}
        \begin{subfigure}{\textwidth}
        \includegraphics[width=\textwidth]{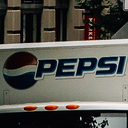} \\ \small HR (img-861)\\ (PSNR$\uparrow$/DISTS$\downarrow$\cite{dists}) \\ TWS$\uparrow$
    \end{subfigure}
    \begin{subfigure}{\textwidth}
        \includegraphics[width=\textwidth]{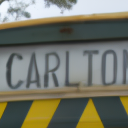} \\ \small LDM-VLM (Ours) \\(27.98 / 0.266) \\ 0.3975
    \end{subfigure}
        \begin{subfigure}{\textwidth}
        \includegraphics[width=\textwidth]{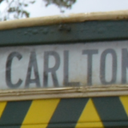} \\ \small HR (img-832)\\ (PSNR$\uparrow$/DISTS$\downarrow$\cite{dists}) \\  TWS$\uparrow$
    \end{subfigure}
\end{subfigure} 
\caption{Visual comparison of the proposed ensembled LDM-VLM method with the state-of-the-art for $\times$4 SR on images from DIV2K validation set~\cite{Agustsson_2017_CVPR_Workshops}. Even though the proposed method has clear advantages in reconstructing realistic high-frequency details while inhibiting artifacts reflected by TWS, the visual improvements are not conveyed by popular quantitative metrics.}
\label{fig:visuals_div2k} 
\end{figure*}

To validate our VLM selection results against human preferences, we conducted a task similar to digit identification, focusing on selecting the~most photorealistic image from a set of 100 diffusion samples for 15 natural images from the DIV2K \cite{Agustsson_2017_CVPR_Workshops} dataset. In each round, participants were asked to select up to 5 images that exhibited the most natural-looking details, colors, and lighting. For both tasks—digit identification and artifact reduction in natural images—no ground truth images were provided, requiring participants to rely solely on their visual perception for decision-making. The Top-5 chosen images are ensembled via pixel-wise averaging and presented in LDM-HumanSelection (LDM-HS) row in Table~\ref{table:fusion_results}. Our observations indicate that the VLMs demonstrated a strong correlation with human preferences, consistently selecting SR images that were generally perceived as more visually appealing and containing fewer artifacts. Furthermore, VLMs offered a scalable and consistent method for assessing SR image quality across large datasets, eliminating the need for extensive human feedback while maintaining a high degree of accuracy in subjective evaluations.




\noindent
\textbf{Qualitative Comparison.} The qualitative results from our experiments highlight the comparative performance of various SR methods, including both GAN-based and diffusion-based approaches. Visual comparisons among 4$\times$ SR approaches and our proposed methods, LDM-VLM (Top-1 and Top-5), are presented in Fig. \ref{fig:visuals_div2k}. Specifically, we observe that state-of-the-art GAN-SR methods like ESRGAN+ \cite{esrganplus}, SPSR \cite{ma_SPSR}, LDL \cite{details_or_artifacts}, and SROOE \cite{srooe_Park_2023_CVPR}, as well as stochastic SR approaches such as HCFLow++ \cite{hcflow_liang21hierarchical} and SRFlow-DA \cite{jo2021srflowda}, often introduce visible artifacts and suffer from excessive sharpness or oversmoothing. Although these methods aim to enhance perceptual quality, they frequently compromise fine details, resulting in artificial textures and distortions that detract from the overall realism of the images. Similarly, random samples without any guidance in diffusion models such as LDM \cite{LDM_rombach2022high} and IDM \cite{implicit_gao2023implicit} lead to unwanted artifacts, particularly around letters and concrete shapes (e.g., PEPSI image). In contrast, our LDM-VLM approach effectively mitigates these distortions by leveraging VLM-guided selection to produce visually cleaner and semantically faithful outputs. Compared to the human feedback-based method (LDM-HS)~\cite{korkmaz2024_icip}, which prioritizes perceptual realism through subjective judgments, LDM-VLM offers an automated and scalable solution that consistently selects samples with fewer artifacts and higher visual clarity.

\begin{figure*}
\centering
\begin{subfigure}{0.16\textwidth}
    \begin{subfigure}{\textwidth}
        \includegraphics[width=\textwidth]{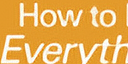} \\ \small HAT \cite{hat_chen2023activating} \\ 0.5917
    \end{subfigure}
    \begin{subfigure}{\textwidth}
        \includegraphics[width=\textwidth]{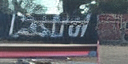} \\ \small HAT \cite{hat_chen2023activating} \\ 0.3354
    \end{subfigure}
\end{subfigure}
\begin{subfigure}{0.16\textwidth}
    \begin{subfigure}{\textwidth}
        \includegraphics[width=\textwidth]{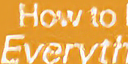} \\ \small  ESRGAN+ \cite{esrganplus} \\ 0.5127
    \end{subfigure}
    \begin{subfigure}{\textwidth}
        \includegraphics[width=\textwidth]{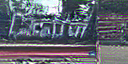} \\ \small  ESRGAN+ \cite{esrganplus} \\ 0.1900
    \end{subfigure}
\end{subfigure}
\begin{subfigure}{0.16\textwidth}
    \begin{subfigure}{\textwidth}
        \includegraphics[width=\textwidth]{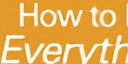} \\\small  SRFlow-DA \cite{jo2021srflowda} \\ 0.6110
    \end{subfigure}
    \begin{subfigure}{\textwidth}
        \includegraphics[width=\textwidth]{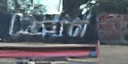} \\\small  SRFlow-DA \cite{jo2021srflowda} \\ 0.3272
    \end{subfigure}
\end{subfigure}
\begin{subfigure}{0.16\textwidth}
    \begin{subfigure}{\textwidth}
        \includegraphics[width=\textwidth]{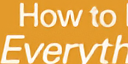} \\ \small  SROOE \cite{srooe_Park_2023_CVPR} \\ 0.6241
    \end{subfigure}
    \begin{subfigure}{\textwidth}
        \includegraphics[width=\textwidth]{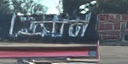} \\ \small SROOE \cite{srooe_Park_2023_CVPR} \\ 0.3239
    \end{subfigure}
\end{subfigure}
\begin{subfigure}{0.16\textwidth}
     \begin{subfigure}{\textwidth}
        \includegraphics[width=\textwidth]{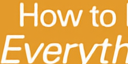} \\ \small LDM-VLM (Ours) \\ 0.6505
    \end{subfigure}
    \begin{subfigure}{\textwidth}
        \includegraphics[width=\textwidth]{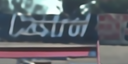} \\ \small LDM-VLM (Ours) \\ 0.3936
    \end{subfigure}
\end{subfigure}
\begin{subfigure}{0.16\textwidth}
    \begin{subfigure}{\textwidth}
        \includegraphics[width=\textwidth]{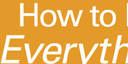} \\ \small HR- Set14\\ TWS$\uparrow$
    \end{subfigure}
        \begin{subfigure}{\textwidth}
        \includegraphics[width=\textwidth]{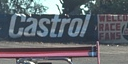} \\ \small HR- BSD100 \\ TWS$\uparrow$
    \end{subfigure}
\end{subfigure} 
\caption{Visual comparison of the proposed ensembled LDM-VLM method with the SoTA for $\times$4 SR on images from SR benchmarks \cite{set14_cite, bsd100_cite}.}
    \label{fig:visuals_benchmark} 
    \end{figure*}

To further validate generalizability, we extend our qualitative evaluation to two additional benchmarks: Set14 and BSD100. As shown in Fig.~\ref{fig:visuals_benchmark}, LDM-VLM continues to outperform competing methods across these datasets, producing sharp, artifact-free results even on challenging natural scenes. These improvements are also reflected in consistently higher TWS values, confirming that our method not only preserves visual appeal but also enhances semantic reliability across a broader range of image types and complexities. While our core focus remains on information-centric scenarios, these results demonstrate the broader applicability and robustness of our approach across standard SR benchmarks.


Overall, both information-centric evaluations and quality assessment of natural images indicate that VLMs offer a scalable and objective approach for selecting diffusion SR samples that closely aligns with human judgment. This positions VLMs as a viable solution for efficiently selecting high-quality, reliable SR outputs without the need for extensive human feedback.

\subsection{Evaluating Image Quality with Vision-Language Models: Correlation with Human Perception}
To assess how well VLMs align with human perception in image quality assessment, we evaluated GPT-4o Mini~\cite{gpt4_achiam2023gpt} on the KADID-10k dataset~\cite{kadid10k}, which includes human-rated degraded images. We selected six pristine images and applied five common distortions—Gaussian blur, JPEG compression, denoised noise~\cite{dncnn_zhang2017beyond}, pixelation, and intensity quantization~\cite{otsu_4310076}—resulting in 150 degraded samples.

GPT-4o Mini was prompted to rate each the quality of the image on a five-point scale, and its scores were compared against human Mean Opinion Scores (MOS). The model’s assessments correlate well with human ratings (e.g., 0.71 for natural scenes), though performance drops on structurally complex images (e.g., 0.5 on repetitive textures), highlighting limitations in fine-grained perception. Overall, GPT-4o Mini demonstrates strong potential for automated quality assessment, particularly when ground truth is unavailable. However, combining VLM-based evaluations with traditional metrics like LPIPS or DISTS may yield more robust results across varied image types.

\begin{table}[t]
\centering
\caption{Prompt robustness and consistency analysis of different VLMs across two axes: digit/letter recognition (Info) and artifact detection (Artifact). Consistency indicates how often a model selected the same SR sample across prompt variants. Human agreement reflects alignment between VLM-selected and human-preferred outputs. 
}
\label{tab:prompt_robustness}
\begin{tabular}{lcccc}
\toprule
\multirow{2}{*}{\textbf{Model}} & \multicolumn{2}{c}{\textbf{Consistency (\%)}} & \multicolumn{2}{c}{\textbf{Human Agreement (\%)}} \\
\cmidrule(r){2-3} \cmidrule(r){4-5}
& Info & Artifact & Info & Artifact \\
\midrule
GPT-4o          & 95.2 & 93.6 & 91.4 & 89.8 \\
GPT-4o-mini     & 93.5 & 91.0 & 89.5 & 86.7 \\
BLIP-2          & 94.1 & 92.4 & 90.8 & 87.9 \\
LLaMA OCR       & 87.6 & 84.9 & 84.5 & 80.7 \\
O1 Model        & 90.8 & 88.2 & 85.9 & 83.6 \\
\bottomrule
\end{tabular}
\end{table}

\subsection{Prompt Robustness Analysis of VLM-Based Evaluation}
We conducted a prompt robustness analysis for our VLM-based SR sample selection framework using benchmark datasets including DIV2K, BSD100, and Urban100.  In this experiment, we curated a diverse set of prompt variants targeting two key aspects of evaluation: (1) information recognition (e.g., ``What is the digit?" vs. ``Which number is visible?") and (2) artifact detection (e.g., ``How realistic does this image look?" vs. ``Is this image clean and artifact-free?"). For each axis, we constructed a pool of 20 semantically equivalent prompts. These were created through a combination of synonym substitution (e.g., ``digit” vs ``number”), syntactic restructuring (e.g., question rephrasing from passive to active voice), and surface-level paraphrasing (e.g., changing ``Does this image contain distortions?" to ``Is the image free of visual errors?" to simulate natural linguistic variation. The full list of prompts is provided in Appendix A.

For each VLM, we generated predictions using all 20 prompt variants per axis and recorded the selected SR sample per image. We define a model as consistent if it selects the same SR sample across all prompts for a given image. 
Additionally, to assess alignment with human evaluations, we compute human agreement based on how often the prompt-ensemble selection by each model matches human selections.

Results in Table~\ref{tab:prompt_robustness} show that leading models like GPT-4o and BLIP-2 exhibit high consistency and strong agreement with human preferences, with prompt sensitivity below 7\%. Lower-performing models, such as LLaMA OCR and O1, are more susceptible to prompt phrasing, especially for ambiguous or artifact-heavy cases. These findings demonstrate that while VLMs are generally robust to prompt phrasing, incorporating a pool of diverse prompts improves interpretability and reduces selection bias. Accordingly, we adopt a prompt-ensemble strategy in our final implementation to ensure more reliable and consistent SR sample selection across varied input conditions.
\vspace{-18pt}

\subsection{Ablation Study on TWS Weighting Strategy}
\label{sec:tws_ablation}
To evaluate the contribution of each component in our proposed TWS we conduct an ablation study on the DIV2K validation set and Table~\ref{tab:tws_ablation} summarizes the results. We first optimize the weights on a representative image (image 45 in Fig.~\ref{fig:first_img}), yielding the configuration $\lambda_{CLIP}=0.2$, $\lambda_{edge}=0.3$, and $\lambda_{wavelet}=0.5$. To verify the generalizability of this setting, we compare it against several alternative configurations including equal weights for all components ($\lambda=1/3$), and exclusion of the semantic term (CLIP), edge-based SSIM and the wavelet term. Our proposed configuration achieves the highest TWS of 0.3613, confirming that the balanced integration of all three terms is critical for accurate trustworthiness estimation. Removing any component—particularly wavelet-based artifact suppression—leads to a noticeable drop in performance. Notably, removing the CLIP-based semantic term results in lower TWS despite minor differences in pixel structure, reinforcing the importance of high-level understanding in trustworthy SR evaluation. These results validate both the effectiveness and robustness of our weighting scheme across natural image samples and further support the use of our proposed configuration in general SR settings.

\begin{table}[t]
\centering
\caption{Ablation study of TWS weight configurations on DIV2K validation set.}
\label{tab:tws_ablation}
\begin{tabular}{lcc}
\toprule
\textbf{Weight Configuration} & \textbf{TWS (↑)} \\
\midrule
$\lambda_{CLIP}=0.2$, $\lambda_{edge}=0.3$, $\lambda_{wavelet}=0.5$ (ours) & \textbf{0.3613} \\
Equal Weights ($\lambda=1/3$ each) & 0.3491 \\
No CLIP ($\lambda_{CLIP}=0$, $\lambda_{edge}=0.4$, $\lambda_{wavelet}=0.6$) & 0.3387 \\
No Edge ($\lambda_{CLIP}=0.3$, $\lambda_{edge}=0$, $\lambda_{wavelet}=0.7$) & 0.3214 \\
No Wavelet ($\lambda_{CLIP}=0.5$, $\lambda_{edge}=0.5$, $\lambda_{wavelet}=0$) & 0.2940 \\
\bottomrule
\end{tabular}
\end{table}

\section{Discussion}
Our framework selects a single trustworthy SR output from diffusion-generated samples, making it especially suitable for information-critical tasks such as digit or character recognition. While some applications may require multiple outputs, our method is scalable and can be extended to provide ranked alternatives when needed. We use a lightweight ensembling strategy that fuses top-ranked samples identified by VLMs, prioritizing semantic accuracy and artifact suppression without the computational cost of full-sample averaging. Unlike traditional ensemble methods, our approach emphasizes information integrity and aligns closely with human judgment—particularly in cases where metrics like PSNR or SSIM fail to capture semantic errors, such as digit misidentification. We chose BLIP-2 and GPT-4o for their advanced visual reasoning and structured prompt capabilities. Additionally, we validated our method across other VLMs, including LLaMA-OCR and the O1 model, observing consistent alignment with human preferences. These results confirm the robustness and generalizability of our approach. 
While our current prompt design relies on semantically diverse, manually curated queries, we acknowledge the need to further evaluate the system’s robustness under adversarial or misleading prompts. As future work, we plan to systematically construct adversarial prompts—both syntactically confusing and semantically ambiguous—to test the limits of VLM consistency and trustworthiness.
Furthermore, we aim to explore lightweight fine-tuning or prompt adaptation mechanisms (e.g., prompt tuning or instruction finetuning) to enhance model robustness in such adversarial scenarios. These improvements would help solidify the deployment-readiness of VLM-guided SR selection across variable or noisy language conditions.
In addition, expanding to cross-lingual prompts and domain-specific contexts (e.g., medical imaging or remote sensing) is an important future direction. 

Overall, our method provides a reliable, cost-efficient solution for SR tasks where semantic correctness is paramount.
\section{Conclusion}
\label{conc}

By combining the interpretive power of vision-language models (VLMs) with diffusion-based SR, we introduce a scalable and fully automated framework for selecting reliable samples. Our approach leverages VLM-guided evaluation to identify and ensemble the most trustworthy outputs, eliminating the need for costly and time-consuming human feedback. This is particularly valuable in large-scale, information-critical applications where semantic accuracy is essential. To quantitatively support sample selection, we propose the Trustworthiness Score (TWS), a hybrid metric that captures semantic similarity, structural consistency, and artifact suppression. Experimental results demonstrate that samples selected via our VLM-based method consistently achieve the highest TWS across benchmarks—often surpassing those chosen by human annotators—highlighting the effectiveness of our strategy in producing perceptually and semantically faithful SR outputs. While VLMs excel in structured and well-defined tasks, their performance can depend heavily on the quality of provided prompts and contextual information. This study lays the foundation for a new direction in trustworthy SR, bridging generative modeling with intelligent sample evaluation.

\appendices
\section{Prompt Pools for VLM-Based SR Evaluation}
\label{appendix:prompt_pool}

We designed 20 
prompts for each evaluation axis to assess the robustness of VLM responses under varied linguistic formulations. These prompts cover variations in wording, structure, and style, while preserving the semantic intent.
\vspace{-4pt}

\subsection{Digit/Letter (Information) Identification Prompts}
\begin{enumerate}
\item What is the digit in this image?
\item Can you identify the number?
\item Which number is shown here?
\item Please read the digit.
\item What number is visible?
\item Can you tell which number appears?
\item Read the digit from the image.
\item Identify the number in this picture.
\item What does the digit look like?
\item What is written in the image?
\item Is there a digit shown here?
\item Recognize the number in this image.
\item What number can you see?
\item What digit does the image contain?
\item Tell me the number you observe.
\item What's the printed number?
\item Do you recognize a digit?
\item Read the numeral in this image.
\item What digit do you detect?
\item State the digit shown in the image.
\end{enumerate}
\vspace{-16pt}

\subsection{Artifact Detection (Visual Quality) Prompts}
\begin{enumerate}
\item Does this image contain visual artifacts?
\item Is the image clean and artifact-free?
\item Can you spot any distortions?
\item Are there imperfections in this image?
\item How clean is the image?
\item Does this image look realistic?
\item Rate the visual clarity of the image.
\item Is the output free of compression artifacts?
\item Do you notice any artifacts?
\item Are there visible distortions or glitches?
\item Comment on the image’s realism.
\item Does the image appear sharp and clear?
\item Is this image blurry or distorted?
\item Does the output seem natural and artifact-free?
\item Are there distracting visual flaws?
\item Is the image degraded in any way?
\item How visually appealing is this image?
\item Are there any unwanted textures or glitches?
\item Would you consider this image clean?
\item Is this result free of visual anomalies?
\end{enumerate}

\section{Statistical Significance Tests}
We conducted a two-sample t-test comparing consistency scores of GPT-4o and GPT-4o-mini across 20 semantically equivalent prompts for identifying the digit ``45.''  Each model was asked to answer 20 different phrasings of a digit recognition question 
and the number of times it correctly predicted ``45'' was recorded for each prompt. The full list of prompts is provided in Appendix A.
We applied an independent two-sample t-test assuming equal variance to the resulting prompt-wise scores from both models. The results yield a statistically significant difference (t-statistic = 3.09, p-value = 0.0034), indicating that GPT-4o significantly outperforms GPT-4o-mini as expected in maintaining consistency across prompts.

To verify prompt robustness, we also conducted a one-sample t-test for each model against the hypothetical chance-level performance. GPT-4o’s results (t-statistic = 3.31, p-value = 0.0037) confirm that it performs significantly above chance, while GPT-4o-mini’s results (t-statistic = -11.56, p-value \textless 0.0001) indicate a strong and statistically significant deviation below that level. This statistical analysis supports the robustness claims made in our prompt sensitivity evaluation and demonstrates that observed differences in model performance are not due to random variation.

\vspace{18pt}
\printbibliography

\begin{IEEEbiography}[{\includegraphics[width=1in,height=1.25in,clip,keepaspectratio]{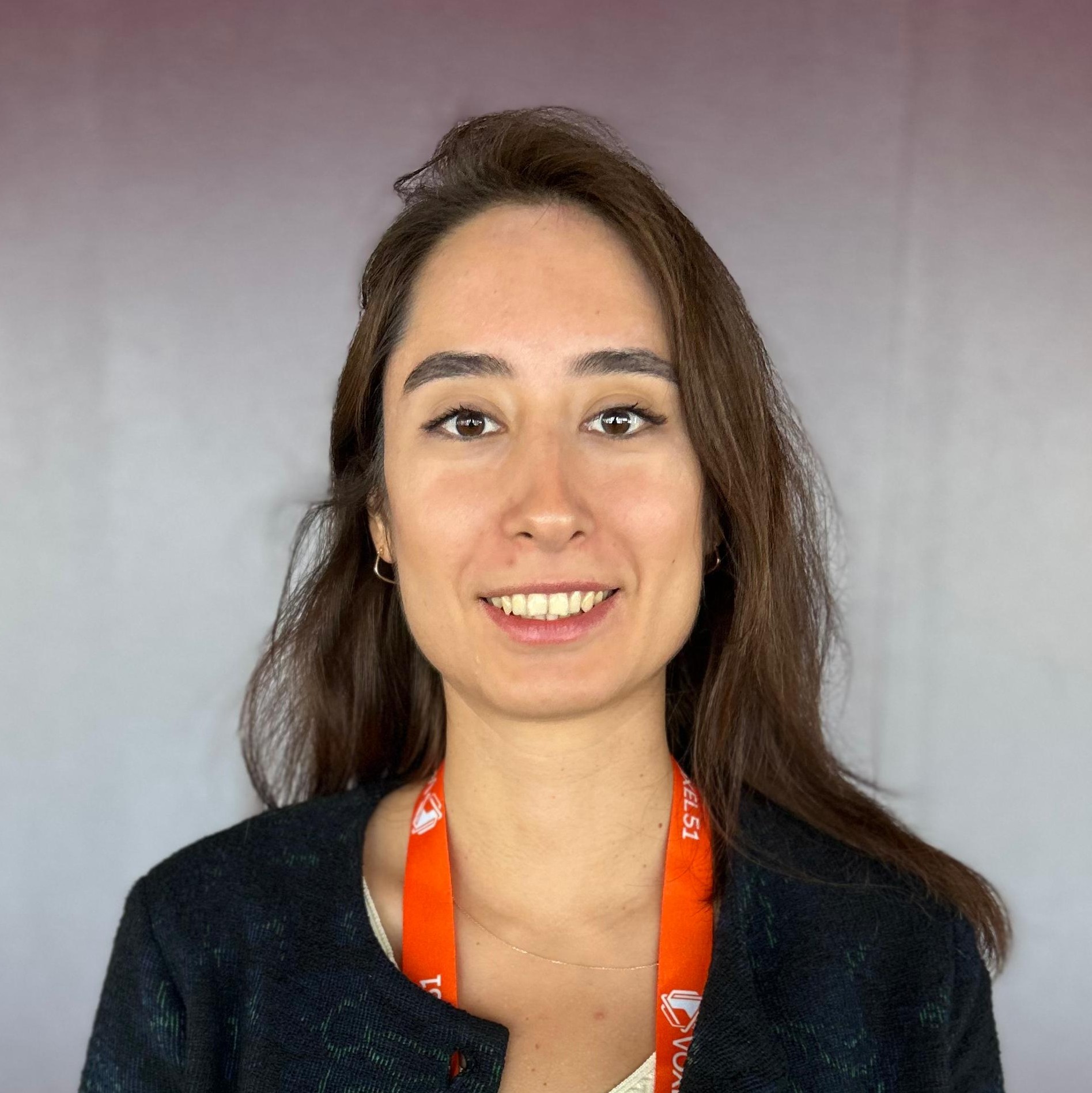}}]
{Cansu Korkmaz} (M'19) received her B.S. and M.S. degrees in Electrical and Electronics Engineering from Koç University in 2019 and 2021, respectively. She is currently pursuing a Ph.D. at Koç University and is affiliated with the KUIS AI Lab. In addition to her Ph.D. studies, she is currently a visiting researcher at the University of Wuerzburg’s Computer Vision Lab, working under the supervision of Prof. Radu Timofte. Her research interests focus on image super-resolution, generative models, computer vision, and deep learning for image processing.
\end{IEEEbiography}

\begin{IEEEbiography}[{\includegraphics[width=1in,height=1.25in,clip,keepaspectratio]{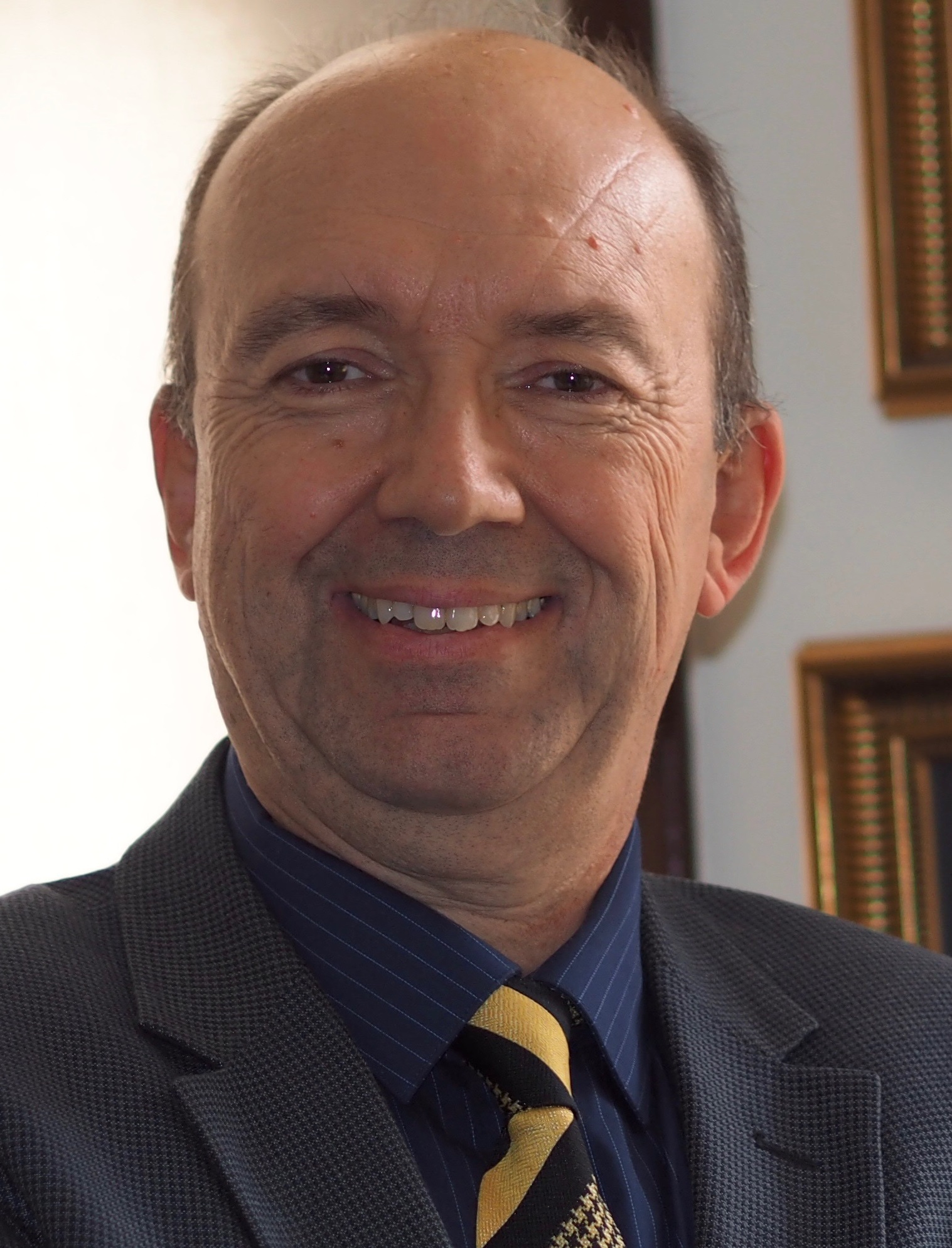}}]{A. Murat Tekalp} (S'80-M'84-SM'91-F'03) received Ph.D. degree in Electrical, Computer, and Systems Engineering from Rensselaer Polytechnic Institute (RPI), Troy, New York, in 1984, He was with Eastman Kodak Company, Rochester, New York, from 1984 to 1987, and with the University of Rochester, Rochester, New York, from 1987 to 2005, where he was promoted to Distinguished University Professor. He is currently Professor at Koc University, Istanbul, Turkey. He served as Dean of Engineering between 2010-2013. His research interests are in digital image and video processing, including video compression and streaming, video networking, multi-view and 3D video processing, and deep learning for image/video processing and compression.

He has been elected a member of Turkish Academy of Sciences and Academia Europaea. He served as an Associate Editor for the IEEE Trans. on Signal Processing (1990-1992) and IEEE Trans. on Image Processing (1994-1996). He was the Editor-in-Chief of the EURASIP journal Signal Processing: Image Communication published by Elsevier between 1999-2010. He was on the Editorial Board of the IEEE Signal Processing Magazine (2007-2010) and the Proceedings of the IEEE (2014-2020). He chaired the IEEE Signal Processing Society Technical Committee on Image and Multidimensional Signal Processing (Jan. 1996 - Dec. 1997). He was appointed as the General Chair of IEEE International Conference on Image Processing (ICIP) at Rochester, NY in 2002. He served in the European Research Council (ERC) Advanced Grant Panels (2009-2015). He is currently in the Editorial Board of Wiley-IEEE Press. He is the Technical Program Co-Chair for IEEE ICIP 2020. Prof. Tekalp has authored the Prentice Hall book Digital Video Processing (1995), a completely rewritten second edition of which is published in 2015. 
\end{IEEEbiography}

\begin{IEEEbiography}[{\includegraphics[width=1in,height=1.25in,clip,keepaspectratio]{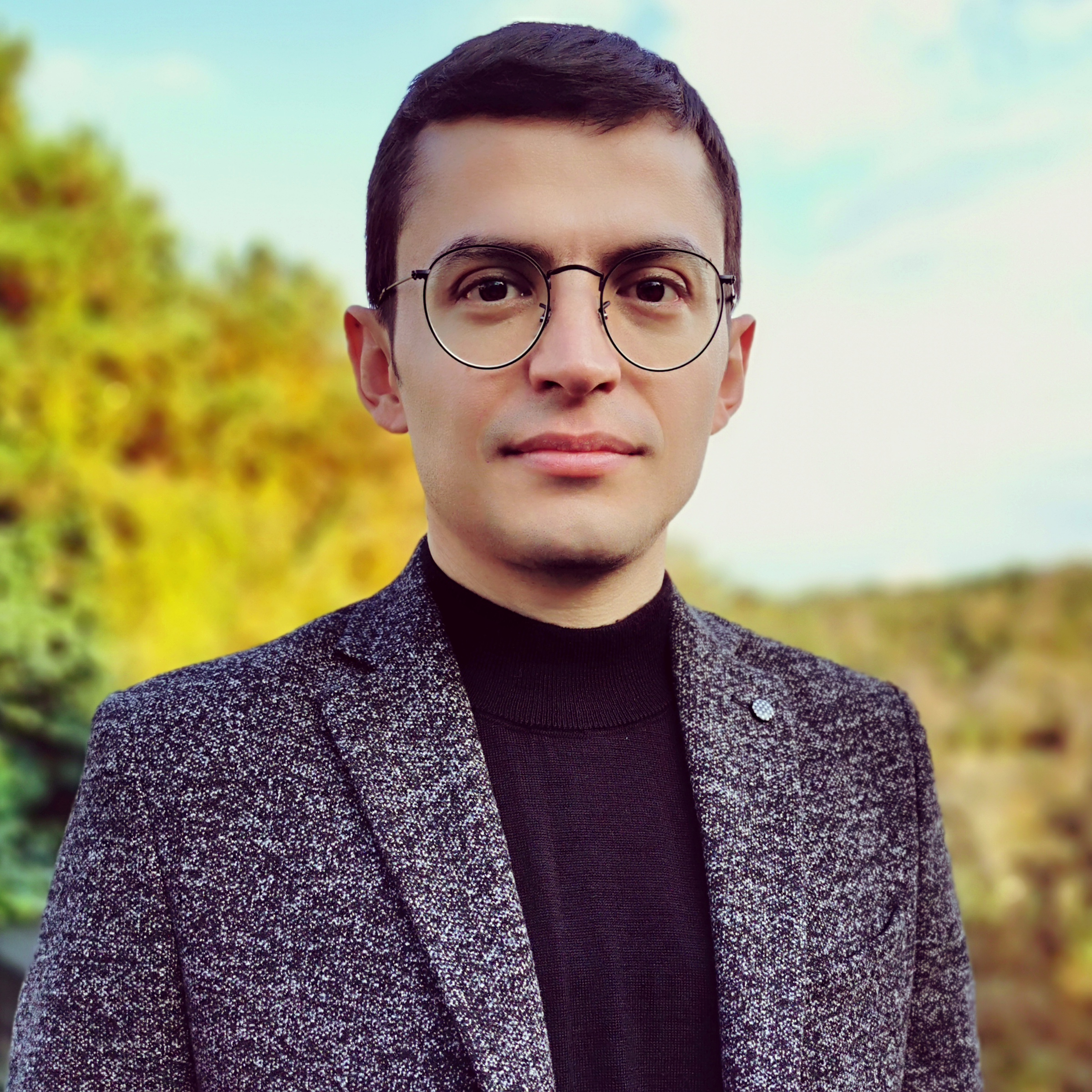}}]{Zafer Do\u{g}an} (M'11) obtained his B.S. in Electrical and Electronics Engineering from Middle East Technical University (METU) in 2009, followed by an M.S. and Ph.D. in Electrical Engineering from École Polytechnique Fédérale de Lausanne (EPFL) in 2011 and 2015, respectively. His Ph.D. research focused on sparse signal representation in data processing and inverse problems in nonlinear models like tomography and neuroimaging. He worked as a postdoctoral research associate at EPFL from 2015 to 2016 and at Harvard University's John A. Paulson School of Engineering and Applied Sciences from 2016 to 2019. Currently, he is an assistant professor in the Department of Electrical and Electronics Engineering at Koç University, leading the Machine Learning and Information Processing (MLIP) Group, and is involved with the KUIS Artificial Intelligence Research Center and the IEEE SPS Turkey Chapter.

His research interests are at the intersection of signal processing, image processing, inverse problems, and machine learning. He currently focuses on the exact dynamics of learning algorithms for large-scale learning and inference problems, interpretability and explainability of artificial learning models, and theoretical understanding of non-convex optimization and deep learning frameworks. Apart from theoretical aspects, he explores specific applications of artificial learning frameworks in image enhancement, computer vision, computational imaging, recommendation systems and autonomous systems to provide enhanced stability, tractability and reproducibility features.
\end{IEEEbiography}





\end{document}